\pdfoutput=1
\documentclass[journal]{IEEEtran}
\usepackage{amsmath,amsfonts}
\usepackage{algorithmic}
\usepackage{algorithm}
\usepackage{array}
\usepackage[caption=false,font=normalsize,labelfont=sf,textfont=sf]{subfig}
\usepackage{textcomp}
\usepackage{stfloats}
\usepackage{url}
\usepackage{verbatim}
\usepackage{graphicx}
\usepackage{cite}
\usepackage{hyperref}
\usepackage{cleveref}
\usepackage{booktabs}
\usepackage{multirow}
\usepackage{mdwlist}
\usepackage{float}
\usepackage{subfig}
\usepackage{color,xcolor}

\hypersetup{
hidelinks,
colorlinks=true,
linkcolor=red,
citecolor=cyan
}

\begin{document}

\title{DS-Depth: Dynamic and Static Depth Estimation via a Fusion Cost Volume}

\author{Xingyu Miao, Yang Bai, Haoran Duan, Yawen Huang, Fan Wan, Xinxing Xu, \\ Yang Long,~\IEEEmembership{Senior Member~IEEE}, Yefeng Zheng,~\IEEEmembership{Fellow~IEEE}
\thanks{Xingyu Miao, Yang Long, Haoran Duan and Fan Wan are with the Department of Computer Science, Durham University(E-mail: \{xingyu.miao; yang.long; fan.wan\}@durham.ac.uk; haoran.duan@ieee.org).}
\thanks{Yang Bai and Xinxing Xu are with the Institute of High Performance Computing (IHPC),
ASTAR, Singapore 138632, Singapore (E-mail: \{bai\_yang; xuxinx\}@ihpc.a-star.edu.sg).}
\thanks{Yawen Huang and Yefeng Zheng are with Tencent Jarvis Lab, Shenzhen, China (E-mail: \{yawenhuang; yefengzheng\}@tencent.com).}
\thanks{Yang Long and Haoran Duan are the corresponding authors.}

\thanks{Manuscript received April, 2023;}}

\markboth{IEEE Transactions on Circuits and Systems for Video Technology}%
{Shell \MakeLowercase{\textit{et al.}}: A Sample Article Using IEEEtran.cls for IEEE Journals}


\maketitle

\begin{abstract}
Self-supervised monocular depth estimation methods typically rely on the reprojection error to capture geometric relationships between successive frames in static environments. However, this assumption does not hold in dynamic objects in scenarios, leading to errors during the view synthesis stage, such as feature mismatch and occlusion, which can significantly reduce the accuracy of the generated depth maps. To address this problem, we propose a novel dynamic cost volume that exploits residual optical flow to describe moving objects, improving incorrectly occluded regions in static cost volumes used in previous work. Nevertheless, the dynamic cost volume inevitably generates extra occlusions and noise, thus we alleviate this by designing a fusion module that makes static and dynamic cost volumes compensate for each other. In other words, occlusion from the static volume is refined by the dynamic volume, and incorrect information from the dynamic volume is eliminated by the static volume. Furthermore, we propose a pyramid distillation loss to reduce photometric error inaccuracy at low resolutions and an adaptive photometric error loss to alleviate the flow direction of the large gradient in the occlusion regions. We conducted extensive experiments on the KITTI and Cityscapes datasets, and the results demonstrate that our model outperforms previously published baselines for self-supervised monocular depth estimation. 
\textit{Code: \url{https://github.com/xingy038/DS-Depth}}.
\end{abstract}


\section{Introduction}
Currently, depth information plays a significant role in several fields, including autonomous vehicles \cite{geiger2012we}, robots \cite{choi2013fast}, AR/VR applications \cite{luo2020consistent}, and 3D reconstructions \cite{newcombe2011real}. Although professional hardware can provide relatively accurate depth information, its high cost precludes widespread use. An alternative approach is to use RGB cameras, which generate a sequence of RGB images that can be leveraged by self-supervised monocular depth prediction methods \cite{garg2016unsupervised, song2021monocular, cao2018monocular, mohaghegh2018aggregation, ye2019deep}. While this approach addresses the expensive cost of professional hardware, its performance still falls short of that of professional hardware or deep multi-view methods. Nevertheless, self-supervised monocular depth prediction methods show promise and are gaining popularity in both research and industrial communities. However, estimating depth from a monocular image is an ill-posed problem due to the numerous plausible depth values that can exist in the same 3D scene, with the depth information able to project countless identical 2D scenes.

\begin{figure}[t]
  \centering
   \includegraphics[width=1\linewidth]{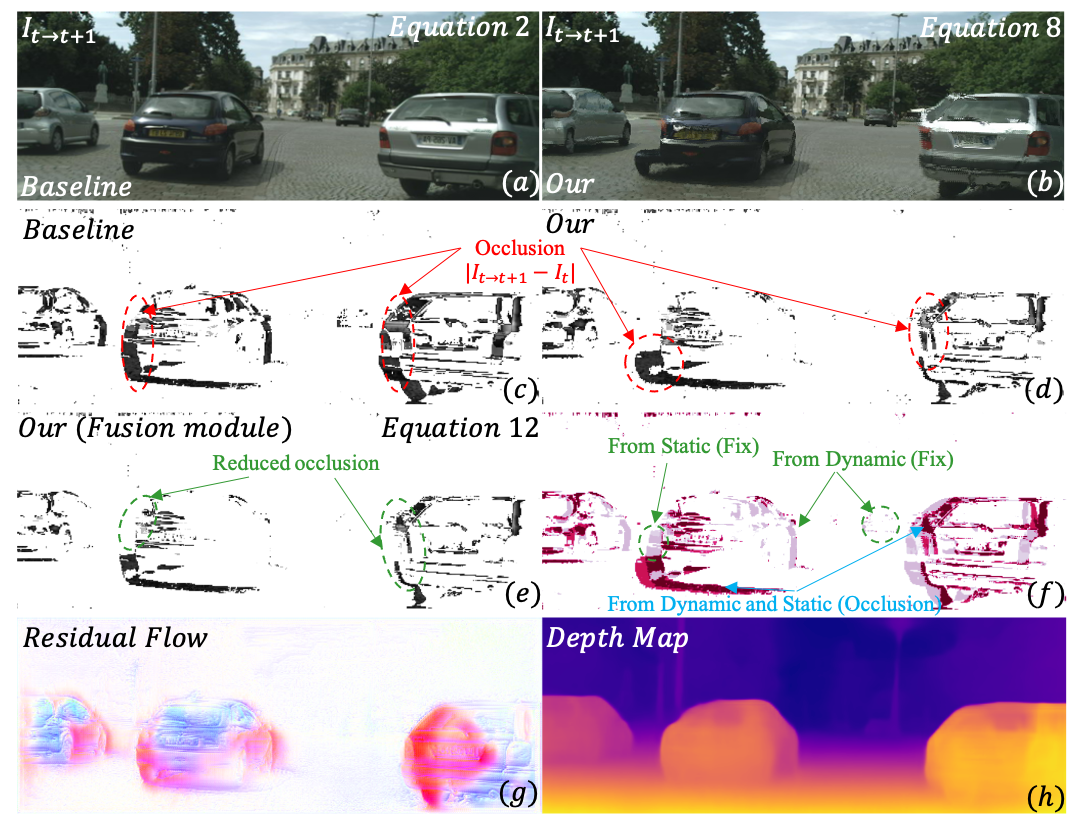}
   \caption{Our baseline \cite{manydepth} is based on a static environment to generate a cost volume. However, this approach is limited when there are dynamic objects in the scene, which can cause errors such as mismatches and occlusions (Figures (a) and (c)). To handle this issue, we propose a novel dynamic cost volume that incorporates residual optical flow (Figure (g)) to capture object motion. While the dynamic cost volume can reduce occlusions, it may also introduce new occlusions and noise (Figures (b) and (d)). To overcome this challenge, we design an adaptive fusion module that makes the static and dynamic cost volumes complement each other. This approach leads to further occlusion reduction and significant improvements in performance (Figures (e) and (f)).}
   \label{fig:0}
\end{figure}

Previous self-supervised depth estimation methods rely on multi-frame information during training but only use the current frame as input during inference \cite{godard2017unsupervised, monodepth2, guizilini20203d, shu2020feature, zhou2017unsupervised, yin2018geonet}. In contrast, multi-frame self-supervised depth estimation methods employ multi-frame information during both training and inference stages, typically by constructing a cost volume \cite{ke2021deep, manydepth, feng2022disentangling, guizilini2022multi, sun2018pwc, long2021multi, wimbauer2021monorec, teed2018deepv2d} or utilizing related layers \cite{hur2020self} to learn additional geometric features for improving the performance of the model. Although multi-frame methods seem to perform better, it heavily relies on feature matching to establish geometric relationships between frames. Therefore, multi-frame methods will fail in some cases, especially encountering moving objects. Currently, both the single-frame method and the multi-frame method use photometric error loss to train the model. This kind of loss is based on the static environment, once there is a moving object between consecutive frames, which will mislead the model and bring some wrong information. Additionally, despite the fact that the existing work on self-unsupervised depth estimation \cite{manydepth,wimbauer2021monorec,guizilini2022multi,guizilini2022learning, meng2021cornet} provides excellent solutions for unlabeled data, they are incapable of effectively handling dynamic objects, which limits their applicability to real-world scenarios.

In this work, we present DS-Depth, a self-supervised depth estimation framework designed to achieve general applicability in depth estimation. To mitigate occlusions caused by dynamic objects during view synthesis, as depicted in \Cref{fig:0}, we incorporate residual optical flow to refine the cost volume generated by ego-motion, resulting in a dynamic cost volume that accurately captures the scene's dynamics. However, it is worth noting that dynamic cost volume alone is insufficient to resolve the occlusion problem while bringing extra occlusion and noise. Therefore, we propose a fusion module to combine the dynamic and static cost volumes and mitigate the occlusion and noise during view synthesis. In order to make the fused cost volume obtain the more correct gradient, we design an adaptive photometric error loss to alleviate the large gradient in the occlusion region. Additionally, during the training process, we introduce a pyramid distillation loss to alleviate the inaccuracies of photometric errors at low resolutions, leading to more accurate predicted depth maps. In summary, the contributions of our work are as follows:
\begin{enumerate}
  \item
  We propose a novel dynamic cost volume using camera ego-motion and residual optical flow to construct, which improves upon the cost volume constructed from camera ego-motion to handle occlusions caused by moving objects.
  \item
  We investigate the occlusion and noise in dynamic cost volume approaches and propose an adaptive fusion module that makes static and dynamic cost volumes compensate for each other. Our experimental results show a significant improvement in the estimated depth.
  \item
  We design a pyramid error loss to improve the photometric error at low resolution and an adaptive photometric error loss to make the fused cost volume get a more accurate gradient. Experiments show that our design is effective.
  \item
  We achieve state-of-the-art depth estimation results on two challenging datasets KITTI and Cityscapes.
\end{enumerate}

\section{Related Works}
\subsection{Self-supervised monocular depth estimation}

More recently, self-unsupervised monocular depth estimation has been a kind of promising method for the limited labeled depth data, which aims to from a single image to predict a pixel-level depth map. The original self-supervised depth estimation framework is proposed by Zhou et al. \cite{zhou2017unsupervised}, which leverages a DpethNet and PostNet to predict the geometry relationship between frames. This framework originally is used for stereo pairs, and then it was extended to the monocular. Additionally, Godard et al.\cite{godard2017unsupervised} consider the depth estimation task as a view synthesis problem and minimize the image reconstruction objective. In terms of the view synthesis, Monodepth2 \cite{monodepth2} propose a minimal reprojection error to address occlusion and reduce visual artifacts using full-resolution multiscale sampling. In terms of the additional losses, FeatDepth \cite{shu2020feature} designs a new reconstruction error metric, which improves depth prediction of the low-texture area. For camera geometry modeling, Gordon et al. \cite{gordon2019depth} first explore learning the camera intrinsic parameters through the network so that the model can be applied in wild videos. In terms of the network architectures, PackNet \cite{guizilini20203d} aims to solve the problem that traditional encoder (such as ResNet) leads to the resolution being reduced and thus losing some details.

\subsection{Multi-frame Monocular Depth Estimation}

Single-frame depth estimation is based on the depth cues, such as motion information, linear perspective, occlusion, texture, and shadow \cite{kumar2018depth, wei2021iterative, lee2016adaptive}. The accuracy of these cues determines the estimated depth \cite{zhou2017unsupervised,godard2017unsupervised,monodepth2,shu2020feature,gordon2019depth,guizilini20203d}. Early multi-frame depth estimation approaches use test-time refinement methods \cite{casser2019,chen2019self,luo2020consistent,mccraith2020monocular,shu2020feature,kuznietsov2021comoda} and recurrent neural networks \cite{cs2018depthnet, patil2020don} to improve the performance of the model. 

The stereo matching method performs feature matching by correcting the image with a known baseline, thus the method has been transformed into a pixel-by-pixel disparity estimation in the horizontal direction \cite{kendall2017end, liang2018learning}. The test-time refinement method employs a monocular approach to use temporal information at test time, while the recurrent neural network combines a monocular depth estimation network to process continuous frame sequences. However, models using recurrent neural networks are often computationally expensive and have no explicit geometric inference method. Multi-view stereo (MVS) matches any number of views \cite{long2021multi,huang2018deepmvs,im2019dpsnet,yao2018mvsnet, kendall2017end, liang2018learning}, however, most methods are supervised, and some recent self-supervised methods exploit the cost volume in stereo matching combined with single-frame features for geometric inference \cite{manydepth,wimbauer2021monorec,feng2022disentangling,guizilini2022multi,hwang2022self}. Similar to MVS, these works first pre-define a set of depths through which the reference frame is warped to the target frame, and then compute the difference between this frame and the target frame, stacking this difference to form a cost volume. In the cost volume space, the hypothesized depth with the lowest value is closest to the true depth. By using cost volume, the performance of the model has been greatly improved, however, these methods also use reprojection error as training loss, the model fails when encountering scenes with dynamic objects. These methods either try to avoid dynamic objects or use a single-frame model as a teacher-guided cost volume to alleviate this problem. We propose to use optical flow to describe dynamic objects to synthesize correct views, thereby building a credible cost volume to improve the performance of the model. 

In addition, we use cost volume to solve the ambiguity of multiple depths, while \cite{chen2021fixing} utilizes the consistency of point clouds for the same purpose. Additionally, they address the matching problem caused by photometric errors, while we tackle the issue of occlusions. \cite{wang20223d} employs a layered method to refine the camera pose and generate a depth map, whereas we apply a similar approach by manipulating the cost volume to adjust inaccurate information and obtain a more precise depth map. Although \cite{jin2018depth} and \cite{wei2021iterative} enhance the quality of the depth map through depth bin and iterative features, respectively, they do not consider the impact of moving objects on the results.

\begin{figure*}
  \centering
   \includegraphics[width=1\linewidth]{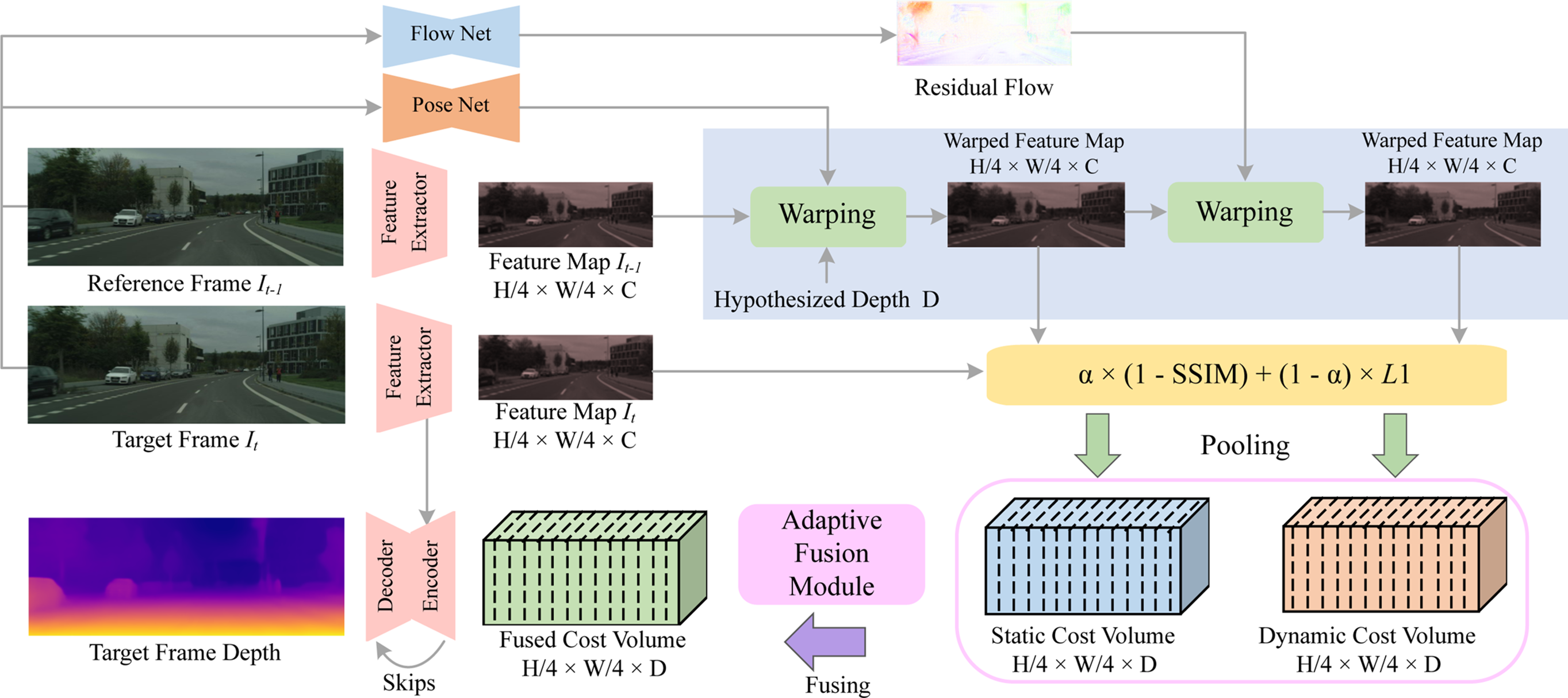}
   \caption{\textbf{The pipeline of our network} is depicted and comprises four main components: a multi-frame depth estimation network, a single-frame depth estimation network, a pose network, and an optical flow module. Specifically, we adopt a ResNet18 \cite{he2016deep} as the backbone. We employ the same pose network as used in \cite{monodepth2}. In contrast to \cite{manydepth}, our model generates two cost volumes, which are fused using a specialized fusion module. Additionally, we generate the dynamic cost volume by combining ego-motion with the residual flow, where the residual flow is obtained from FlowNet\cite{yin2018geonet}.}
   \label{fig:4}
\end{figure*}

\subsection{Dynamic Objects in Self-supervised Depth Prediction}
Currently, due to self-supervised depth prediction using reprojection error is not suitable for dynamic scenes. Thus, the key to solving this problem is how to separate static objects and dynamic objects. Additionally, separating static environments and dynamic objects can also improve the robustness of depth estimation in dynamic situations. There have been some works\cite{gordon2019depth,manydepth,feng2022disentangling,wimbauer2021monorec,klingner2020self,lee2021learning,lee2021attentive} that propose solutions to train static and moving objects separately, which aim to solve the problem that moving objects in assumed static environment cannot be reprojected well to the correction position. 

Although, above all methods have some good performance, there are still some limitations. For example, the input frame is a single frame that cannot clarify the temporal information, and disentangling moving objects and static objects will increase the complexity of the model.  In addition, for methods using cost volume, dynamic object regions can mislead the model to build the wrong cost volume, this cost volume with negative information will affect the gradient of the model, causing the model to incorrectly learn the motion region. When the model finally predicts depth, it mistakenly thinks that the motion region is infinite, which reduces the accuracy of the model and affects the performance of the model. Although the method using the cost volume is better than the single-frame method, it still needs to consider the influence of moving objects. 

In addition, several approaches have been proposed for jointly learning optical flow and depth using multi-task networks\cite{yang2018every, hur2020self, hur2021self, liu2019unsupervised, yin2018geonet}, where the optical flow can be indirectly recovered for capturing dynamic objects. While our method uses the residual optical flow to fix the reprojection inaccurate regions in the cost volume.

\section{Method}
\subsection{Preliminaries}
In this section, we briefly introduce the use of view synthesis to construct similar images and then compute their correlations to construct cost volumes.

\subsubsection{View Synthesis}
Following \cite{monodepth2,manydepth}, we use view synthesis as supervision signals. Let two frames $I_{t-1}$ and $I_t$ from an input video as the source image and target image where pixel from target image can be expressed as:
\begin{equation}
z p_{s\to t} =  K[R|T]_{t\to s}D_t K^{-1} p_s
  \label{eq:1}
\end{equation}
where $K$ denotes the camera intrinsic and $[R|T]$ is the camera extrinsic, $p$ denotes the 2D pixel coordinates, and we use the estimated depth map of the target image, relative pose from the target image to the source image, and source image to synthesize target image.

\subsubsection{Cost Volume Construction}
Based on the \Cref{eq:1}, which can be transformed into view synthesis between two features $F_t$ and $F_{t+i}$ in temporal. Thus, The synthesized frame $\hat F_{t+i}$ can be expressed as:
\begin{equation}
\hat F_{t+i} = F_t\left \langle proj(D_t, T_{t\to t+i}, K) \right \rangle .
  \label{eq:2}
\end{equation}
$F\left \langle \cdot \right \rangle$ is bilinear sampling operation and $proj$ is the resulting 2D coordinates of the projected depths $D_t$, which is equal \Cref{eq:1}. Note that when building the cost volume, the predicted depth is replaced by a predefined set of depth values ($D_t = \{D_1,\dots, D_{i-1}, D_i, i\in [1, N]\}$) to generate a synthetic voxel. Leveraging this synthetic voxel ($V^s$) to calculate the correlation (The commonly used methods are SSIM \cite{wang2004image}, difference of absolute values and dot product.) with the target frame ($I_t$), and the final correlation cost volume ($CV$) is obtained. As we describe in \Cref{sec:4.1}, a cost volume constructed in this way can produce erroneous information in dynamic scenarios. Therefore, we introduce residual optical flow to refine the moving object information in cost volume.

\subsection{Architecture}
Previous works \cite{manydepth,wimbauer2021monorec} construct the cost volume of two successive frames solely based on camera ego-motion learned from pose networks, which is less capable of handling dynamic regions/objects as explained in \Cref{sec:4.1}.
To address this problem, we propose dynamic cost volumes, which can effectively handle the dynamic regions/objects from the static cost volume, while also bringing some noise. Thus, based on the static and dynamic cost volumes, we carefully design an adaptive fusion module to handle static and dynamic scenarios pixel-wise to alleviate this case. The overview of our DS-Depth is shown in \Cref{fig:4}.

\subsection{Dynamic Cost Volume Construction}
\label{sec:4.1}
The differences that need to be captured when constructing the cost of two frames essentially originate from 1) the motion of moving objects and 2) changes in the relative camera pose between the two frames. This difference creates occlusions that affect the performance of the model. To better model the effect of moving objects, which was ignored in previous cost volume based on depth estimation networks \cite{manydepth}, we first define the relationship between two-frame correspondences in a 3D scene as follows.

There is a 3D point $X$ in the space, the 2D point projected by $X$ in the frame $I$ at time $t$ is $u$. We define $X^{sen}_{t\to t+1}$ as the motion of the $X$ from time $t$ to time $t+1$ in the 3D scene. When using a known intrinsic camera to observe point $X_t$, we define $\mathcal{P}(X_t)$ as the projection of the $X_t$ to the image coordinate $u_t$. Additionally, we define the ego-motion of the camera as $u^{cam}_{t\to t+1}$, and the move of optical flow as $u^{opt}_{t\to t+1}$.

\begin{figure}[t]
\centering
\includegraphics[width=0.9\linewidth]{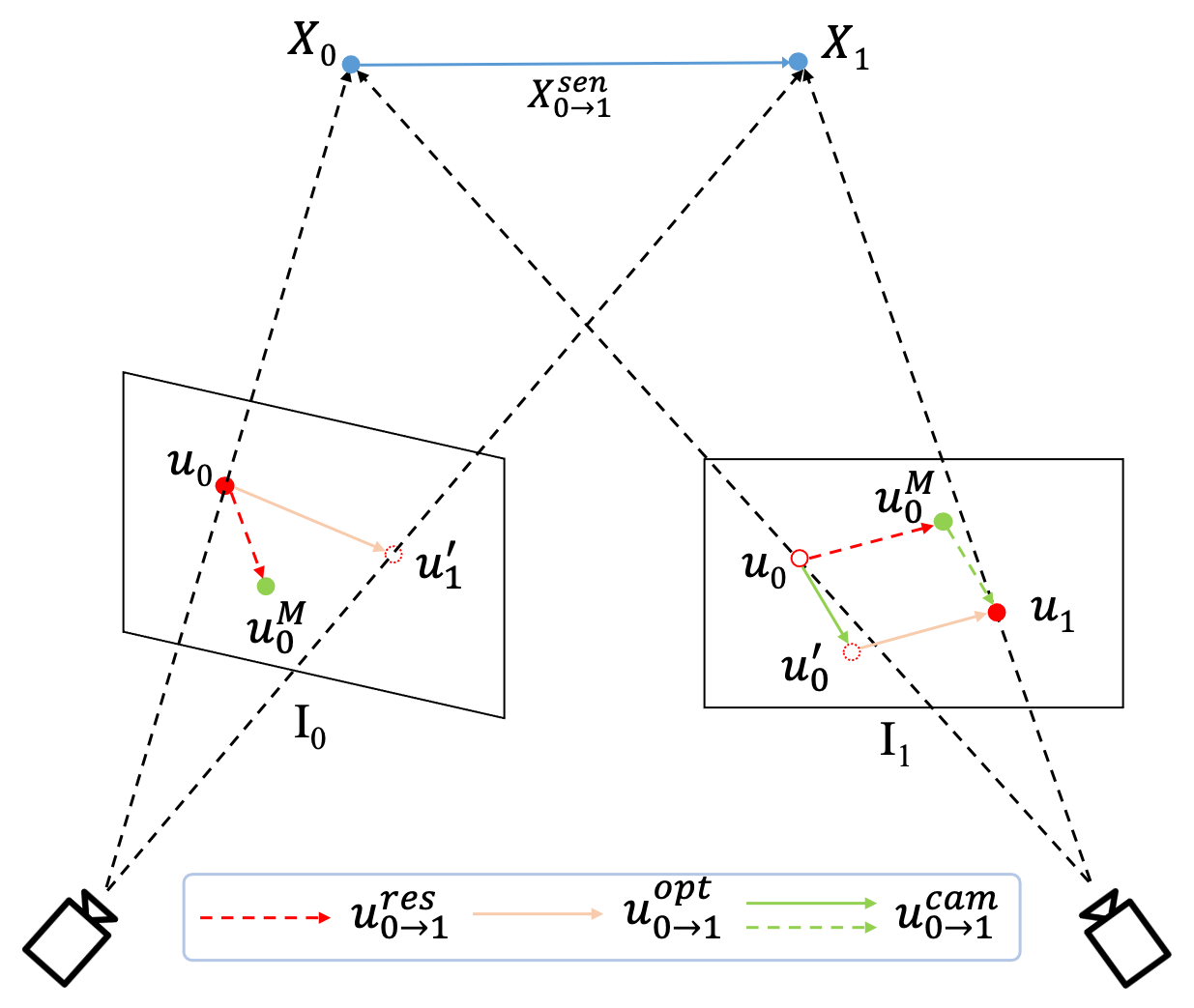}
\caption{\textbf{Geometric relationship of points on two frames}. 
Consider two frames, denoted $I_0$ and $I_1$, where the camera moves from $I_0$ to $I_1$. Let $X_0$ and $X_1$ be two points in 3D space, and let their corresponding 2D coordinates be projected onto the frames as $u_0$ and $u_1$, respectively. Note that $u_0^\prime$ is the projection of $X_0$ onto $I_1$, and if $X_0$ is observed in $I_1$, its location can be calculated by reprojecting the camera, and its motion can be expressed as $u^{cam}_{0\to 1}$. If the camera is static and observes $X_0$ and $X_1$ at the same time, the motion from $X_0$ to $X_1$ can be expressed as $X^{sen}_{0\to 1}$. Then, the motion of the 2D coordinates $u_0^\prime$ and $u_1$ on $I_1$ can be expressed as a projected optical flow, denoted $u^{opt_1}_{0\to 1}$. Similarly, on $I_0$, this projected optical flow can be expressed as $u^{opt_0}_{0\to 1}$. Finally, $u^{res_0}_{0\to 1}$ represents the residual flow learned by our optical flow module, and $u_0^M$ is obtained by moving $u_0$ by residual optical flow.}
\label{fig:2}
\end{figure}

As shown in \Cref{fig:2}, for ease of understanding here, we set $t$ and $t+1$ to 0 and 1, respectively. The projection of optical flow in a 3D perspective can be expressed as:
\begin{equation}
u_{0\to 1}^{opt} = \mathcal{P}(T_1(X_0+X_{0\to1}^{sen}))-\mathcal{P}(T_0X_0)
  \label{eq:4}
\end{equation}

where $T$ denotes the known camera extrinsic matrix for $I_t$. Intuitively, \Cref{eq:4} could represent projected scene flow $X_{0\to1}^{sen}$ on 2D plane. However, since the camera is moving, the view motion should be taken into account. To be specific, the 2D correspondences moves $I_0$ to $I_1$ can be expressed as:
\begin{equation}
u_{0\to 1}^{cam} = \mathcal{P}(T_1X_0)-\mathcal{P}(T_0X_0)
  \label{eq:5}
\end{equation}

Thus, the $u_0$ observed on $I_0$ should be represented as $u_0^\prime$ on $I_1$, thereby scene flow $X_{0\to1}^{sen}$ can be derived as:
\begin{equation}
X_{0\to1}^{sen} = T^{-1}_{1}\mathcal{P}^{-1}(u_0^\prime+u_{0\to1}^{opt_1},D_1)-T^{-1}_{0}\mathcal{P}^{-1}(u_0^\prime,D_0)
  \label{eq:6}
\end{equation}
where $D$ indicates $i_{th}$ depth level. Based on \Cref{eq:6}, we need to know two different depths ($D_0$ and $D_1$) to represent the scene flow $X_{0\to1}^{sen}$ in space, but in our framework, we can only predict $D_0$. Therefore, we use an optical flow module to learn the residual flow $u_{0\to1}^{res_0}$ on the frame $I_0$ combined with the camera ego-motion to represent the scene flow. 
As shown in \Cref{fig:2}, the 2D point $u_0$ move to $u_1$ can be expressed as:
\begin{equation}
(u_0\to u_1) = u_{0\to 1}^{cam} + u_{0\to 1}^{opt_1}
  \label{eq:14}
\end{equation}

However, $u_{0\to 1}^{opt1}$ is not equal to $u_{0\to 1}^{opt0}$ because the camera is moving, which causes the corresponding motion of $u_0$. Therefore, we utilize an optical flow network to learn a residual flow $u_{0\to1}^{res_0}$ that approximates $u_{0\to 1}^{opt1}$. In this case, we can substitute $u_{0\to 1}^{opt1}$ with $u_{0\to1}^{res_0}$, resulting in the following relationship:
\begin{equation}
(u_0\to u_1) = u_{0\to 1}^{cam} + u_{0\to1}^{res_0}
  \label{eq:15}
\end{equation}
thus, similar to \Cref{eq:2}, new synthesized feature $\hat F_t^N$ is:
\begin{equation}
\hat F_{t+i}^N = \hat F_t(\left \langle proj(D_t, T_{t\to t+i}, K) \right \rangle + u_{t\to t+i}^{res_t}).
  \label{eq:13}
\end{equation}

To construct dynamic cost volume, we follow a similar approach to Manydepth, but with the added consideration of dynamic object movement. Specifically, when generating a target view from a reference view, we incorporate residual optical flow and camera transformation information into the synthesis process. Furthermore, we have enhanced the construction of our cost volume by utilizing a projection error consistent method. While Manydepth employs an $L_1$ approach, we adopt a formulation that combines structural similarity index measure (SSIM) and $L_1$, which is:
\begin{equation}
\mathcal{E}(\hat F,F) = \alpha (1-\text{SSIM} (\hat F,F))+(1-\alpha)\left \| \hat F-F \right \| _1
\label{eq:18}
\end{equation}
where $\alpha$ is 0.4. The yielded cost volume is expressed as:
\begin{equation}
CV_S=\mathcal{E}(\hat F,F)
\label{eq:11}
\end{equation}
\begin{equation}
CV_D=\mathcal{E}(\hat F^N,F)
\label{eq:12}
\end{equation}
where $CV_S$ is the static cost volume that is constructed with camera movement, and $CV_D$ is the dynamic cost volume that uses residual optical flow to refine static cost volume, feature map $\hat F$ is generated by \Cref{eq:2} and feature map $\hat F^N$ is generated by \Cref{eq:13}.

\begin{table*}
\caption{\textbf{Depth estimation results} on KITTI and Cityscapes. The methods in this table are classified according to multi-frame and single-frame, Abs.Rel. error descending order, where the best method is in \textbf{bold} and the second best is \underline{underlined}. Additionally, the table only shows the results of splitting the KITTI with Eigen et. al. \cite{eigen2015predicting} and the Cityscapes split with \cite{zhou2017unsupervised}. \\
\textbf{K} – KITTI\quad \textbf{C} – Cityscapes}
\label{tab:1}
\begin{tabular}{ccccccccccc}
\toprule[1pt]
\multirow{2}{*}{Method}                                            & \multirow{2}{*}{Test frames} & \multirow{2}{*}{Dateset} & \multirow{2}{*}{WxH} & \multicolumn{4}{c}{Lower is better} & \multicolumn{3}{c}{Higher is better}                    \\ \cline{5-11} 
                                                                   &                              &                          &                      & AbsRel  & SqRel  & RMSE   & RMSElog & $\delta < 1.25$ & $\delta < 1.25^2$ & $\delta < 1.25^3$ \\ \hline
Ranjan et al. \cite{ranjan2019competitive}        & 1                            & K                        & 832x256              & 0.148   & 1.149  & 5.464  & 0.226   & 0.815           & 0.935             & 0.973             \\
EPC++ \cite{luo2019every}                         & 1                            & K                        & 832x256              & 0.141   & 1.029  & 5.350  & 0.216   & 0.816           & 0.941             & 0.976             \\
Struct2depth (M) \cite{casser2019}                & 1                            & K                        & 416x128              & 0.141   & 1.026  & 5.291  & 0.215   & 0.816           & 0.945             & 0.979             \\
Li et al. \cite{li2020unsupervised}               & 1                            & K                        & 416x128              & 0.130   & 0.950  & 5.138  & 0.209   & 0.843           & 0.948             & 0.978             \\
Videos in the wild \cite{gordon2019depth}         & 1                            & K                        & 416x128              & 0.128   & 0.959  & 5.230  & 0.212   & 0.845           & 0.947             & 0.976             \\
Monodepth2 \cite{monodepth2}                      & 1                            & K                        & 640x192              & 0.115   & 0.903  & 4.863  & 0.193   & 0.877           & 0.959             & 0.981             \\
Packnet-SFM \cite{guizilini20203d}                & 1                            & K                        & 640x192              & 0.111   & 0.785  & 4.601  & 0.189   & 0.878           & 0.960             & 0.982             \\
Johnston et al. \cite{johnston2020self}           & 1                            & K                        & 640x192              & 0.106   & 0.861  & 4.699  & 0.185   & 0.889           & 0.962             & 0.982             \\
Guizilini et al. \cite{guizilini2020semantically} & 1                            & K                        & 640x192              & 0.102   & \textbf{0.698}  & \underline{4.381}  & 0.178   & 0.896           & 0.964             & 0.984             \\
Patil et al. \cite{patil2020don}                  & N                            & K                        & 640x192              & 0.111   & 0.821  & 4.650  & 0.187   & 0.883           & 0.961             & 0.982             \\
Wang et al. \cite{wang2020self}                   & 2 (-1,0)                     & K                        & 640x192              & 0.106   & 0.799  & 4.662  & 0.187   & 0.889           & 0.961             & 0.982             \\
ManyDepth \cite{manydepth}                        & 2 (-1,0)                     & K                        & 640x192              & \underline{0.098}   & \underline{0.770}  & 4.459  & \underline{0.176}   & \underline{0.900}          & \underline{0.965}             & \underline{0.983}             \\
\textbf{Our}                                                                & 2 (-1,0)                     & K                        & 640x192              & \textbf{0.095}   & \textbf{0.698}  & \textbf{4.329}  & \textbf{0.173}   & \textbf{0.905}           & \textbf{0.966}             & \textbf{0.984}             \\ \hline
Pilzer et al. \cite{pilzer2018unsupervised}       & 1                            & C                        & 512x256              & 0.240   & 4.264  & 8.049  & 0.334   & 0.710           & 0.871             & 0.937             \\
Struct2depth 2 \cite{casser2019}                  & 1                            & C                        & 416x128              & 0.145   & 1.737  & 7.280  & 0.205   & 0.813           & 0.942             & 0.976             \\
Monodepth2 \cite{monodepth2}                      & 1                            & C                        & 416x128              & 0.129   & 1.569  & 6.876  & 0.187   & 0.849           & 0.957             & 0.983             \\
Videos in the wild \cite{gordon2019depth}         & 1                            & C                        & 416x128              & 0.127   & 1.330  & 6.960  & 0.195   & 0.830           & 0.947             & 0.981             \\
Li et al. \cite{li2020unsupervised}               & 1                            & C                        & 416x128              & 0.119   & 1.290  & 6.980  & 0.190   & 0.846           & 0.952             & 0.982             \\
Lee et al. \cite{lee2021attentive}                & 1                            & C                        & 832x256              & 0.116   & 1.213  & 6.695  & 0.186   & 0.852           & 0.951             & 0.982             \\
Struct2Depth 2 \cite{casser2019}                  & 3 (-1,0,1)                   & C                        & 416x128              & 0.151   & 2.492  & 7.024  & 0.202   & 0.826           & 0.937             & 0.972             \\
ManyDepth \cite{manydepth}                        & 2 (-1,0)                     & C                        & 416x128              & \underline{0.114}   & \underline{1.193}  & \underline{6.223}  & \underline{0.170}   & \underline{0.875}           & \underline{0.967}             & \underline{0.989}             \\
\textbf{Our}                                                                & 2 (-1,0)                     & C                        & 416x128              & \textbf{0.100}   & \textbf{1.055}  & \textbf{5.884}  & \textbf{0.155}  & \textbf{0.899}           & \textbf{0.974}             & \textbf{0.991}             \\ \bottomrule[1pt]
\end{tabular}
\end{table*}

\subsection{Adaptive Fusion Module}
\label{sec:4.3}
\begin{figure}[!ht]
\centering
\includegraphics[width=1\linewidth]{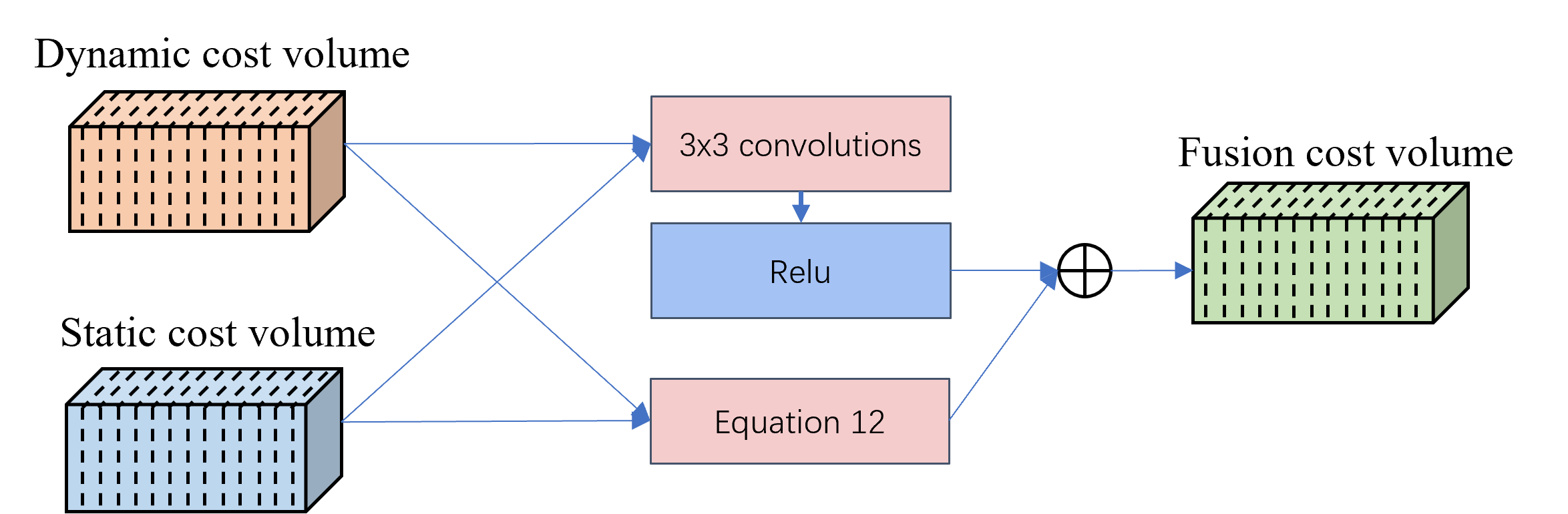}
\caption{\textbf{Adaptive Optical Flow Fusion Module}. 
Here, we describe our adaptive optical flow fusion module, which consists of two branches. The cost volume is a 4D tensor ($B\times D\times \frac{H}{4} \times \frac{W}{4}$). Where $D$ is the depth level, using 96 depth bins.}
\label{fig:flowmodule}
\end{figure}

The cost volume construction involves leveraging the time $t$ feature and the warped time $t-1$ feature. Specifically, the feature $F^w$ at time $t-1$ is warped to time $t$ along a hypothetical depth $D$, after which \Cref{eq:11} is employed to generate the cost volume. However, this process will occur occlusion, which can pollute the cost volume distribution. To address this issue, we employ a learned residual optical flow to simulate object motion and rectify the incorrectly warped pixels caused by dynamic objects, thereby guiding the gradient flow to the correct pixels. Although we use residual optical flow to refine the incorrect information in static cost volume caused by dynamic objects, it will inevitably cause some extra occlusions and noise (As shown in \Cref{fig:0}). Thus, we carefully design a fusion module to alleviate this problem. This fusion module is divided into two branches, one of the branches is a simple concatenate operation, while another branch utilizes static and dynamic cost quantities to adaptive complement each other. The adaptive fusion branch can be expressed as:
\begin{equation}
CV_{com}=
\begin{cases}
 CV_D, &(F^w_o\in CV_S) \cup (F^w_v\in CV_D)   \\
 CV_S, &(F^w_o\in CV_D) \cup (F^w_v\in CV_S)   \\
 min(CV_S, CV_D), & F^w_v\in (CV_{D} \cup CV_{S})
\end{cases}
\label{eq:7}
\end{equation}
where $O/V$ are the set of occluded/visible areas in $F^w$. Specifically, in cases where a pixel in the static cost volume is occluded, we substitute it with the corresponding pixel in the dynamic cost volume and vice versa. Thus, the final fused cost volume is:
\begin{equation}
CV_{f}=CV_{com} + CV_{cat}
\label{eq:16}
\end{equation}
where $CV_{cat}$ is obtained by concatenating the two cost quantities and passing through a simple convolution layer. After experiments, the fusion cost volume can effectively alleviate the partial occlusions and noise problem. The effectiveness of our design is confirmed in ablation studies. The module architecture is shown in \Cref{fig:flowmodule}.

\begin{table*}
\centering
\caption{\textbf{Depth estimation results} on KITTI dataset with improved ground truth\cite{uhrig2017sparsity}. We evaluated our method using the KITTI dataset with improved ground truth and followed convention by sorting methods in each category by their Absolute Relative error with respect to the ground truth. The best methods were highlighted in \textbf{Bold}. Our method surpassed all other state-of-the-art approaches, including some stereo-based and supervised methods. \\
\textbf{Sup} – Supervised by ground truth depth\quad \textbf{S} – Stereo\quad \textbf{M} – Monocular}
\label{tab:4}
\begin{tabular}{cccccccccc}
\toprule[1pt]
\multirow{2}{*}{Method}    & \multirow{2}{*}{Training} & \multirow{2}{*}{WxH} & \multicolumn{4}{c}{The lower the better}                                                          & \multicolumn{3}{c}{The higher the better}                                                        \\ \cline{4-10} 
                           &                           &                      & \multicolumn{1}{c}{Abs Rel} & \multicolumn{1}{c}{Sq Rel} & \multicolumn{1}{c}{RMSE}  & RMSE log & \multicolumn{1}{c}{$\delta < 1.25$} & \multicolumn{1}{c}{$\delta < 1.25^2$} & $\delta < 1.25^3$ \\ \hline
Zhan FullNYU \cite{zhan2018unsupervised}        & Sup                       & 608 x 160            & \multicolumn{1}{c}{0.130}   & \multicolumn{1}{c}{1.520}  & \multicolumn{1}{c}{5.184} & 0.205    & \multicolumn{1}{c}{0.859}           & \multicolumn{1}{c}{0.955}             & 0.981             \\
Kuznietsov et al. \cite{kuznietsov2017semi} & Sup                       & 621 x 187            & \multicolumn{1}{c}{0.089}   & \multicolumn{1}{c}{0.478}  & \multicolumn{1}{c}{3.610} & 0.138    & \multicolumn{1}{c}{0.906}           & \multicolumn{1}{c}{0.980}             & 0.995             \\
DORN \cite{fu2018deep}               & Sup                       & 513 x 385            & \multicolumn{1}{c}{0.072}   & \multicolumn{1}{c}{0.307}  & \multicolumn{1}{c}{2.727} & 0.120    & \multicolumn{1}{c}{0.932}           & \multicolumn{1}{c}{0.984}             & 0.995             \\ \hline
Monodepth \cite{godard2017unsupervised}          & S                         & 512 x 256            & \multicolumn{1}{c}{0.109}   & \multicolumn{1}{c}{0.811}  & \multicolumn{1}{c}{4.568} & 0.166    & \multicolumn{1}{c}{0.877}           & \multicolumn{1}{c}{0.967}             & 0.988             \\
3net \cite{poggi2018learning} (VGG)        & S                         & 512 x 256            & \multicolumn{1}{c}{0.119}   & \multicolumn{1}{c}{0.920}  & \multicolumn{1}{c}{4.824} & 0.182    & \multicolumn{1}{c}{0.856}           & \multicolumn{1}{c}{0.957}             & 0.985             \\
3net \cite{poggi2018learning} (ResNet 50)  & S                         & 512 x 256            & \multicolumn{1}{c}{0.102}   & \multicolumn{1}{c}{0.675}  & \multicolumn{1}{c}{4.293} & 0.159    & \multicolumn{1}{c}{0.881}           & \multicolumn{1}{c}{0.969}             & 0.991             \\
SuperDepth \cite{pillai2019superdepth}        & S                         & 1024 x 384           & \multicolumn{1}{c}{0.090}   & \multicolumn{1}{c}{0.542}  & \multicolumn{1}{c}{3.967} & 0.144    & \multicolumn{1}{c}{0.901}           & \multicolumn{1}{c}{0.976}             & 0.993             \\
Monodepth2 \cite{monodepth2}        & S                         & 640 x 192            & \multicolumn{1}{c}{0.085}   & \multicolumn{1}{c}{0.537}  & \multicolumn{1}{c}{3.868} & 0.139    & \multicolumn{1}{c}{0.912}           & \multicolumn{1}{c}{0.979}             & 0.993             \\
EPC++ \cite{luo2019every}             & S                         & 832 x 256            & \multicolumn{1}{c}{0.123}   & \multicolumn{1}{c}{0.754}  & \multicolumn{1}{c}{4.453} & 0.172    & \multicolumn{1}{c}{0.863}           & \multicolumn{1}{c}{0.964}             & 0.989             \\ \hline
SfMLearner \cite{zhou2017unsupervised}        & M                         & 416 x 128            & \multicolumn{1}{c}{0.176}   & \multicolumn{1}{c}{1.532}  & \multicolumn{1}{c}{6.129} & 0.244    & \multicolumn{1}{c}{0.758}           & \multicolumn{1}{c}{0.921}             & 0.971             \\
Vid2Depth \cite{mahjourian2018unsupervised}         & M                         & 416 x 128            & \multicolumn{1}{c}{0.134}   & \multicolumn{1}{c}{0.983}  & \multicolumn{1}{c}{5.501} & 0.203    & \multicolumn{1}{c}{0.827}           & \multicolumn{1}{c}{0.944}             & 0.981             \\
GeoNet \cite{yin2018geonet}            & M                         & 416 x 128            & \multicolumn{1}{c}{0.132}   & \multicolumn{1}{c}{0.994}  & \multicolumn{1}{c}{5.240} & 0.193    & \multicolumn{1}{c}{0.833}           & \multicolumn{1}{c}{0.953}             & 0.985             \\
DDVO \cite{wang2018learning}              & M                         & 416 x 128            & \multicolumn{1}{c}{0.126}   & \multicolumn{1}{c}{0.866}  & \multicolumn{1}{c}{4.932} & 0.185    & \multicolumn{1}{c}{0.851}           & \multicolumn{1}{c}{0.958}             & 0.986             \\
Ranjan \cite{ranjan2019competitive}            & M                         & 832 x 256            & \multicolumn{1}{c}{0.123}   & \multicolumn{1}{c}{0.881}  & \multicolumn{1}{c}{4.834} & 0.181    & \multicolumn{1}{c}{0.860}           & \multicolumn{1}{c}{0.959}             & 0.985             \\
EPC++ \cite{luo2019every}             & M                         & 832 x 256            & \multicolumn{1}{c}{0.120}   & \multicolumn{1}{c}{0.789}  & \multicolumn{1}{c}{4.755} & 0.177    & \multicolumn{1}{c}{0.856}           & \multicolumn{1}{c}{0.961}             & 0.987             \\
Johnston et al. \cite{johnston2020self}      & M                         & 640 x 192            & \multicolumn{1}{c}{0.081}   & \multicolumn{1}{c}{0.484}  & \multicolumn{1}{c}{3.716} & 0.126    & \multicolumn{1}{c}{0.927}           & \multicolumn{1}{c}{0.985}             & 0.996             \\
Monodepth2 \cite{monodepth2}        & M                         & 640 x 192            & \multicolumn{1}{c}{0.090}   & \multicolumn{1}{c}{0.545}  & \multicolumn{1}{c}{3.942} & 0.137    & \multicolumn{1}{c}{0.914}           & \multicolumn{1}{c}{0.983}             & 0.995             \\
Packnet-SFM \cite{guizilini20203d}       & M                         & 640 x 192            & \multicolumn{1}{c}{0.078}   & \multicolumn{1}{c}{0.420}  & \multicolumn{1}{c}{3.485} & 0.121    & \multicolumn{1}{c}{0.931}           & \multicolumn{1}{c}{0.986}             & 0.996             \\
Patil et al.\cite{patil2020don}       & M                         & 640 x 192            & \multicolumn{1}{c}{0.087}   & \multicolumn{1}{c}{0.495}  & \multicolumn{1}{c}{3.775} & 0.133    & \multicolumn{1}{c}{0.917}           & \multicolumn{1}{c}{0.983}             & 0.995             \\
Wang et al.\cite{wang2020self}        & M                         & 640 x 192            & \multicolumn{1}{c}{0.082}   & \multicolumn{1}{c}{0.462}  & \multicolumn{1}{c}{3.739} & 0.127    & \multicolumn{1}{c}{0.923}           & \multicolumn{1}{c}{0.984}             & 0.996             \\
ManyDepth \cite{manydepth}         & M                         & 640 x 192            & \multicolumn{1}{c}{0.070}   & \multicolumn{1}{c}{0.399}  & \multicolumn{1}{c}{3.455} & 0.113    & \multicolumn{1}{c}{0.941}           & \multicolumn{1}{c}{0.989}             & 0.997             \\
\textbf{Our}                        & M                         & 640 x 192            & \multicolumn{1}{c}{\textbf{0.067}}   & \multicolumn{1}{c}{\textbf{0.359}}  & \multicolumn{1}{c}{\textbf{3.314}} & \textbf{0.109}    & \multicolumn{1}{c}{\textbf{0.946}}           & \multicolumn{1}{c}{\textbf{0.989}}             & \textbf{0.997}           \\ \bottomrule[1pt]
\end{tabular}
\end{table*}

\subsection{Loss Function}
We train our self-unsupervised monocular depth architecture using only the photometric reprojection loss, which includes two parts, a structure similarity (SSIM) \cite{wang2004image} and absolute error (L1) terms:
\begin{equation}
L_p = \frac{\alpha}{2} (1-\text{SSIM} (I_t,\hat{I}_t))+(1-\alpha)\left \| I_t-\hat{I}_t \right \| _1
\label{eq:8}
\end{equation}

Based on previous work we set $\alpha=0.85$. In order to train the residual optical flow, we have changed the synthesis of $\hat{I}_t$. In the previous work, the synthesis of view is via \Cref{eq:2}. Now we first obtain the residual optical flow and then use \Cref{eq:13} for the final view synthesis. We can obtain the new synthetic view $\hat{I}_t^f$. Although the new synthetic view captures the correct dynamic region, it may also introduce additional noise, which can cause larger gradients and ultimately lead to degraded module performance if the view is used directly without any preprocessing or regularization. Therefore, we design an adaptive photometric loss to alleviate this problem:
\begin{equation}
L_p = \frac{\alpha}{2} \mathcal{S} (I_t,\hat{I}_t,\hat{I}_t^f)+(1-\alpha)\mathcal{L}(I_t,\hat{I}_t,\hat{I}_t^f)
\label{eq:08}
\end{equation}
where $\mathcal{S}$ can be expressed as:
\begin{equation}
\mathcal{S} = 1-max(\text{SSIM} (I_t,\hat{I}_t),\text{SSIM} (I_t,\hat{I}_t^f))
\label{eq:09}
\end{equation}
and $\mathcal{L}$ can be expressed as:
\begin{equation}
\mathcal{L} = min(\left \| I_t-\hat{I}_t \right \| _1,\left \| I_t-\hat{I}_t^f \right \| _1)
\label{eq:090}
\end{equation}

We also use edge-aware smoothness for depth regularization:
\begin{equation}
L_{s}=\left|\partial_{x} \hat{d}_{t}\right| e^{-\left|\partial_{x} I_{t}\right|}+\left|\partial_{y} \hat{d}_{t}\right| e^{-\left|\partial_{y} I_{t}\right|}
\label{eq:9}
\end{equation}

Moreover, the photometric consistency measurement is not accurate for low resolution \cite{jonschkowski2020matters, chen2021fixing}. Direct unsupervised training at intermediate levels is not suitable, especially at low resolutions. In this case, we use the last output depth map as a pseudo-label to add supervised learning to the low-resolution output. We directly upsample the lower resolution output and evaluate its difference from the final output. Therefore, our pyramid distillation loss is:
\begin{equation}
L_d = \sum_{i=0}^{N} \frac{1}{HW}\mathcal{F}(D_f-S_\uparrow (s_i,D_f))
\label{eq:10}
\end{equation}
where $i$ is scale, $S_\uparrow$ is upsample operate, and $\mathcal{F}$ is the robust penalty function\cite{liu2019ddflow}: $\mathcal{F}=(|x|+\varepsilon)^q$, $q$, $\varepsilon$ being 0.4 and 0.1.
Finally, our training loss is $L = L_p+L_s+L_{consistency}+L_d$, where $L_{consistency}$ is consistency loss from \cite{manydepth} with no additional modifications.

\section{Experimental Results}
We evaluate our DS-Depth model on two challenging depth estimation datasets (KITTI \cite{geiger2012we} and Cityscapes\cite{cordts2016cityscapes}) and show SOTA results by comparison. Finally, the effectiveness of our model is verified by ablation experiments.

\subsection{Datasets and Experimental Settings}
\subsubsection{KITTI}
The KITTI is a widely-used dataset and is the standard benchmark for depth evaluation. We use Eigen et al. split form \cite{eigen2015predicting} with filtered static frames form Zhou et al.\cite{zhou2017unsupervised}. This segmentation method is mostly used for single-frame depth estimation, but it has also been used for multi-frame depth estimation recently\cite{manydepth,feng2022disentangling}. It includes 39,810 training images, 4,424 validation images, and 697 test images.

\subsubsection{Cityscapes}
The cityscapes contains 150,000 images. Following \cite{yang2018lego,zhou2017unsupervised,manydepth}, we train on 69,731 images, which are split according to how the script in \cite{zhou2017unsupervised}  is split. We do not use any stereo image pairs or semantics. We evaluate our model on the 1,525 test images provided by SGM\cite{hirschmuller2007stereo}.

\begin{table*}
\centering
\caption{\textbf{Ablation study on KITTI.} Evaluate our dynamic cost volume, fusion module, adaptive photometric loss (APM loss) and Pyramid distillation loss (PD loss) on KITTI.}
\label{tab:2}
\begin{tabular}{cccccccccc}
\toprule[1pt]
                                     &                                       & \multicolumn{2}{c}{Fusion module} &                    &       & \multicolumn{4}{c}{Lower is better}                                                                                       \\ \cline{3-4} \cline{7-10} 
\multirow{-2}{*}{Static cost volume} & \multirow{-2}{*}{Dynamic cost volume} & Complementary    & Concatenate    & \multirow{-2}{*}{APM loss} & \multirow{-2}{*}{PD loss} & AbsRel                       & SqRel                        & RMSE                         & RMSElog                      \\ \hline
                  \checkmark                   &                                       &                  &            &    &                           & 0.101                        & 0.784                        & 4.559                        & 0.183                        \\
                                     &       \checkmark                                 &                  &         &       &                           & 0.102                        & 0.761                        & 4.557                        & 0.180                        \\
                          \checkmark            &        \checkmark                                &     \checkmark              &                &  &                           & 0.102                        & 0.775                        & 4.630                        & 0.182                        \\
                              \checkmark        &    \checkmark                                    &                  &               \checkmark &  &                           & 0.101                        & 0.757                        & 4.500                        & 0.178                        \\
                          \checkmark            &                                       \checkmark &                  \checkmark &                \checkmark  &  &                           & 0.096                        & 0.714                        & 4.398                        & 0.174                        \\
                              \checkmark        &   \checkmark                                     &  \checkmark                 &            \checkmark  & \checkmark  &                            & 0.095 & 0.705 & 4.360 & 0.173 \\
                              
                              \checkmark        &   \checkmark                                     &  \checkmark                 &            \checkmark  &  \checkmark &   \checkmark                         & \textbf{0.095} & \textbf{0.698} & \textbf{4.329} & \textbf{0.173} \\\bottomrule[1pt]
\end{tabular}
\end{table*}

\begin{table*}
\centering
   \caption{\textbf{Ablation study on Cityscapes.} Here we demonstrate the effectiveness of our method on the Cityscapes dataset by evaluating it. Our approach leads to substantial improvements on Cityscapes, where a larger number of moving objects are present in both the training and test footage, compared to KITTI.}
   \label{tab:5}
\begin{tabular}{cccccccccc}
\toprule[1pt]
                                     &                                       & \multicolumn{2}{c}{Fusion module} &                 &          & \multicolumn{4}{c}{Lower is better}                                                                                       \\ \cline{3-4} \cline{7-10} 
\multirow{-2}{*}{Static cost volume} & \multirow{-2}{*}{Dynamic cost volume} & Complementary    & Concatenate &\multirow{-2}{*}{APM loss}    & \multirow{-2}{*}{PD loss} & AbsRel                       & SqRel                        & RMSE                         & RMSElog                      \\ \hline
                  \checkmark                   &                                       &                  &                &                      &     & 0.114                        & 1.193                       & 6.226                        & 0.170                        \\
                                     &       \checkmark                                 &                  &                &     &                      
 & 0.109                        & 1.170                        & 6.130                        & 0.162                        \\
                          \checkmark            &        \checkmark                                &     \checkmark              &           &     &                           & 0.104                        & 1.159                        & 6.012                        & 0.159                        \\
                              \checkmark        &    \checkmark                                    &                  &               \checkmark  &   &                        & 0.103                        & 1.137                        & 6.001                        & 0.158                        \\
                              \checkmark        &    \checkmark                                    &                \checkmark  &               \checkmark  &   &                        & 0.102                        & 1.140                        & 5.940                        & 0.156                        \\
                          \checkmark            &                                       \checkmark &                  \checkmark &                \checkmark &  \checkmark  &                       & 0.101                        & \textbf{1.051}                        & \textbf{5.883}                        & 0.156                        \\
                              \checkmark        &   \checkmark                                     &  \checkmark                 &            \checkmark     &   \checkmark                     &  \checkmark  & \textbf{0.100} & 1.055 & 5.884 & \textbf{0.155} \\ \bottomrule[1pt]
\end{tabular}
\end{table*}


\subsubsection{Evaluation Metrics} Following the state-of-the-art methods\cite{manydepth,feng2022disentangling}, we use Absolute Relative Error (Abs Rel), Squared Relative Error (Sq Rel), Root Mean Squared Error (RMSE), Root Mean Squared Log Error (RMSElog), and $\delta_1$, $\delta_2$, $\delta_3$ as the metrics to evaluate the performance of our model.

\subsubsection{Model Parameters and Inference Time} We present the refined parameters of our improved encoder model along with the corresponding inference times in \Cref{tab:6}. Despite the increase in parameters by 0.31M and the inference time ranging from 0.011 to 0.015 seconds per batch, the performance of our model has significantly improved on both databases, particularly on Cityscapes.

\begin{table}
\centering
\caption{\textbf{The model parameters and running time} (unit: s/batch size) of the baseline and our method.}
\label{tab:6}
\begin{tabular}{ccc}
\hline
Methods   & Parameter & Inference time      \\ \hline
Manydepth\cite{manydepth} & 13.64M    & 0.020$\sim$0.034 s/b \\
Our       & 13.95M    & 0.035$\sim$0.045 s/b \\ \hline
\end{tabular}
\end{table}

\subsubsection{Implementation Details}
Our model is implemented using PyTorch and trained on a single NVIDIA RTX3090 GPU. We adopt ResNet18 \cite{he2016deep}, which is pretrained on the ImageNet dataset \cite{deng2009imagenet}, as our backbone. To optimize our model, we use the Adam optimizer \cite{kingma2014adam} with an initial learning rate of 1e-4 for 30 epochs, a batch size of 12, and we reduce the learning rate by a factor of 10 every 10 epochs when training on the KITTI dataset. Following \cite{manydepth}, we freeze the pose and single-frame teacher network for the last 5 epochs. To build the cost volume, we only use the frame $t-1$, and to calculate the loss, we use the $t-1$ and $t+1$ frames. For the Cityscapes dataset, we use a batch size of 8 and freeze network on the 5th epoch.

\subsection{Comparison to State-of-the-art}
\subsubsection{Results on KITTI}
In \Cref{tab:1} we compare our method with other methods, e.g. single-frame methods \cite{monodepth2,guizilini20203d,casser2019}, multi-frame methods \cite{manydepth,wang2020self,patil2020don} and dynamic region optimization method \cite{lee2021attentive}. Our method focuses on dynamic region optimization, however, there are fewer moving objects in this database, and most of them are static scenarios. Thus, our methods and \cite{casser2019,lee2021learning,lee2021attentive,gao2020attentional,gordon2019depth,li2020unsupervised,feng2022disentangling} have minor improvement for this database (Where \cite{feng2022disentangling} uses mask generated by a pretrained segmentation network, i.e. the predicted depth is closely related to the performance of this segmentation network. Furthermore, data splitting is different from most multi-frame methods, hence we did not compare with it.). Moreover, compared to our baseline, our SqRel error improves 9.35\% which means our method predicts fewer depths with large errors (i.e. smaller errors in dynamic regions/objects.). The Abs.Rel. error statistics per pixel in \Cref{fig:6} also confirm that our method is better than multi-frame methods, the number of the Abs.Rel. error per pixel of our method in the interval $[0,0.5]$ is much larger than that of the baseline.

\subsubsection{KITTI benchmark scores}
The original Eigen \cite{eigen2015predicting} split of the KITTI \cite{menze2015object} dataset employs re-projected single-frame raw LIDAR points as the ground truth for evaluation. However, it may contain outliers such as reflections on transparent objects. Thus, we reported results using the original ground truth since it is widely used.

Recently, Jonas et al. \cite{uhrig2017sparsity} introduced a set of high-quality ground truth depth maps for the KITTI dataset. They used a denser ground truth depth map obtained by accumulating 5 consecutive frames and removing the outliers. This improved ground truth depth is provided for 652 of the 697 test frames in the Eigen test split \cite{eigen2015predicting}. In this study, we evaluated our method using these 652 improved ground truth frames and compared the results with existing state-of-the-art published methods in \Cref{tab:4}. To adhere to convention, we clipped the predicted depths to 80 meters to match the Eigen evaluation.

Our method was ranked by the Absolute Relative Error and outperformed all existing state-of-the-art methods, including some stereo-based and supervised methods.

\begin{figure*}[!ht]
	\centering
	\subfloat{
	    \begin{minipage}[t]{0.192\textwidth}
			\centering
			{Input data}
			\includegraphics[width=1\textwidth]{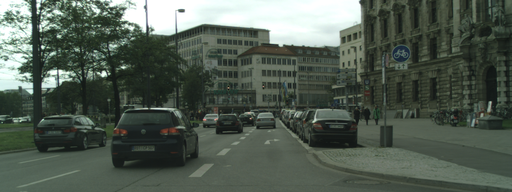}
		\end{minipage}\hspace{-0.5mm}
		\begin{minipage}[t]{0.192\textwidth}
			\centering
			{\cite{manydepth}}\vspace{0.2mm}
			\includegraphics[width=1\textwidth]{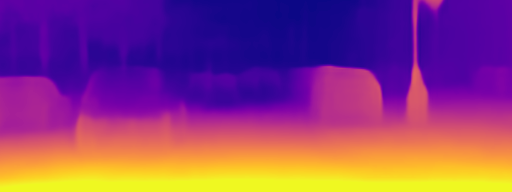}
		\end{minipage}\hspace{-0.5mm}
		\begin{minipage}[t]{0.192\textwidth}
			\centering
			{Error map}
			\includegraphics[width=1\textwidth]{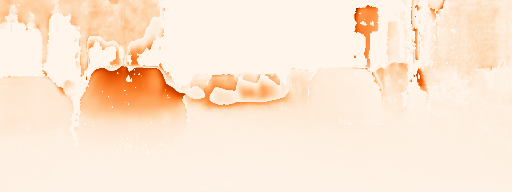}
		\end{minipage}\hspace{-0.5mm}
		\begin{minipage}[t]{0.192\textwidth}
			\centering
			{Our}\vspace{0.8mm}
			\includegraphics[width=1\textwidth]{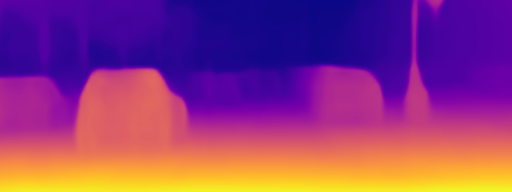}
		\end{minipage}\hspace{-0.5mm}
		\begin{minipage}[t]{0.192\textwidth}
			\centering
			{Error map}
			\includegraphics[width=1\textwidth]{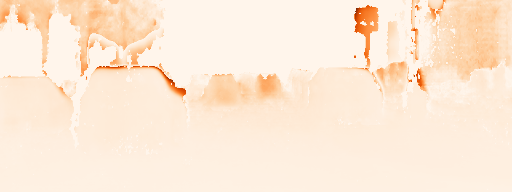}
		\end{minipage}
	}\\\vspace{-3mm}

 	\centering
	\subfloat{
	    \begin{minipage}[t]{0.192\textwidth}
			\centering
			\includegraphics[width=1\textwidth]{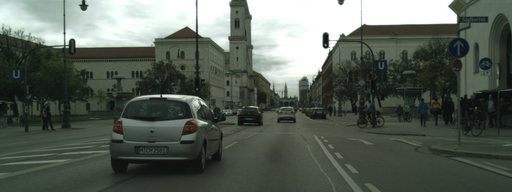}
		\end{minipage}\hspace{-0.5mm}
		\begin{minipage}[t]{0.192\textwidth}
			\centering
			\includegraphics[width=1\textwidth]{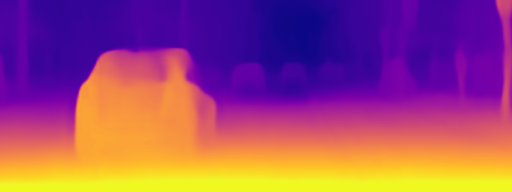}
		\end{minipage}\hspace{-0.5mm}
		\begin{minipage}[t]{0.192\textwidth}
			\centering
			\includegraphics[width=1\textwidth]{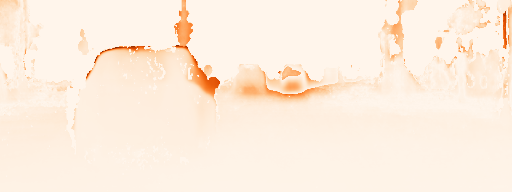}
		\end{minipage}\hspace{-0.5mm}
		\begin{minipage}[t]{0.192\textwidth}
			\centering
			\includegraphics[width=1\textwidth]{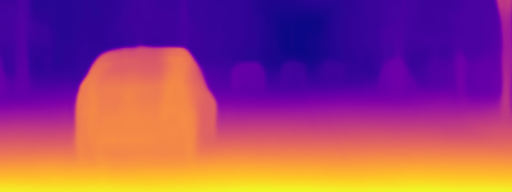}
		\end{minipage}\hspace{-0.5mm}
		\begin{minipage}[t]{0.192\textwidth}
			\centering
			\includegraphics[width=1\textwidth]{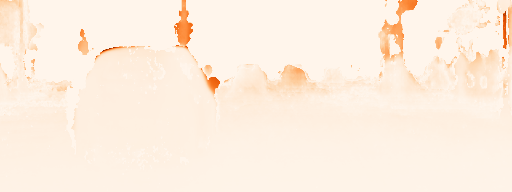}
		\end{minipage}
	}\\\vspace{-3mm}
	
	\centering
	\subfloat{
	    \begin{minipage}[t]{0.192\textwidth}
			\centering
			\includegraphics[width=1\textwidth]{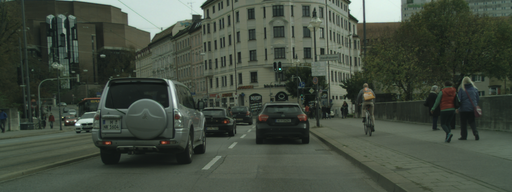}
		\end{minipage}\hspace{-0.5mm}
		\begin{minipage}[t]{0.192\textwidth}
			\centering
			\includegraphics[width=1\textwidth]{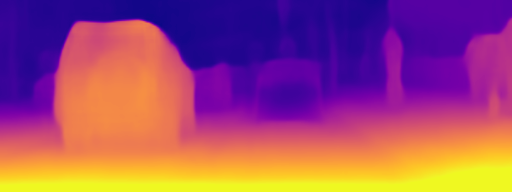}
		\end{minipage}\hspace{-0.5mm}
		\begin{minipage}[t]{0.192\textwidth}
			\centering
			\includegraphics[width=1\textwidth]{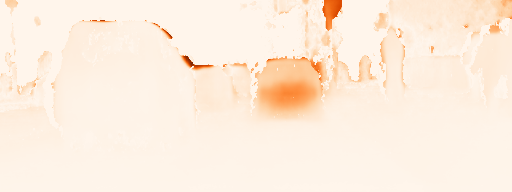}
		\end{minipage}\hspace{-0.5mm}
		\begin{minipage}[t]{0.192\textwidth}
			\centering
			\includegraphics[width=1\textwidth]{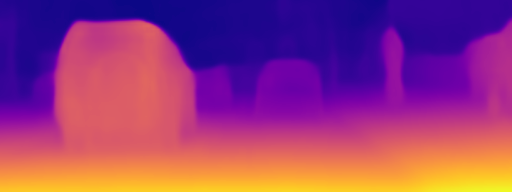}
		\end{minipage}\hspace{-0.5mm}
		\begin{minipage}[t]{0.192\textwidth}
			\centering
			\includegraphics[width=1\textwidth]{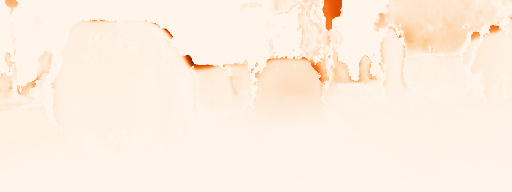}
		\end{minipage}
	}\\\vspace{-3mm}
	
	\centering
	\subfloat{
	    \begin{minipage}[t]{0.192\textwidth}
			\centering
			\includegraphics[width=1\textwidth]{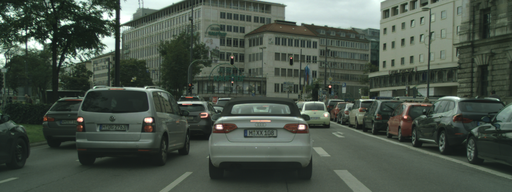}
		\end{minipage}\hspace{-0.5mm}
		\begin{minipage}[t]{0.192\textwidth}
			\centering
			\includegraphics[width=1\textwidth]{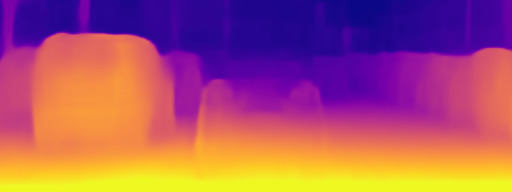}
		\end{minipage}\hspace{-0.5mm}
		\begin{minipage}[t]{0.192\textwidth}
			\centering
			\includegraphics[width=1\textwidth]{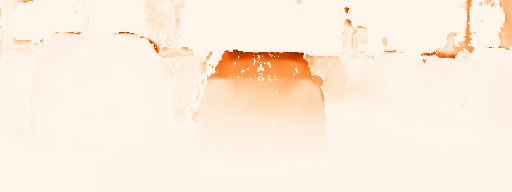}
		\end{minipage}\hspace{-0.5mm}
		\begin{minipage}[t]{0.192\textwidth}
			\centering
			\includegraphics[width=1\textwidth]{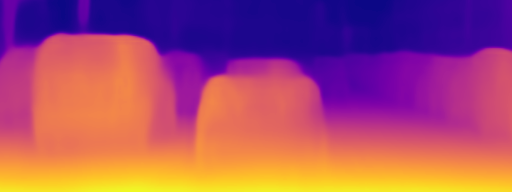}
		\end{minipage}\hspace{-0.5mm}
		\begin{minipage}[t]{0.192\textwidth}
			\centering
			\includegraphics[width=1\textwidth]{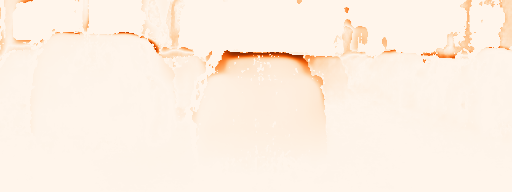}
		\end{minipage}
	}\\\vspace{-3mm}
	
	\centering
	\subfloat{
	    \begin{minipage}[t]{0.192\textwidth}
			\centering
			\includegraphics[width=1\textwidth]{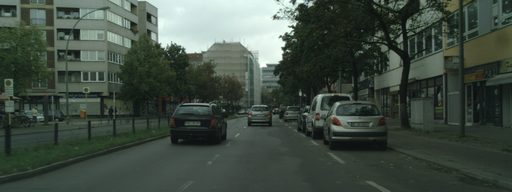}
		\end{minipage}\hspace{-0.5mm}
		\begin{minipage}[t]{0.192\textwidth}
			\centering
			\includegraphics[width=1\textwidth]{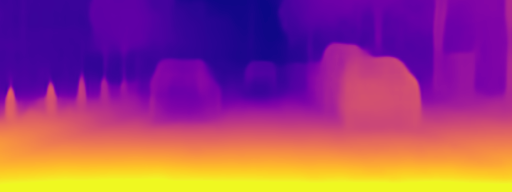}
		\end{minipage}\hspace{-0.5mm}
		\begin{minipage}[t]{0.192\textwidth}
			\centering
			\includegraphics[width=1\textwidth]{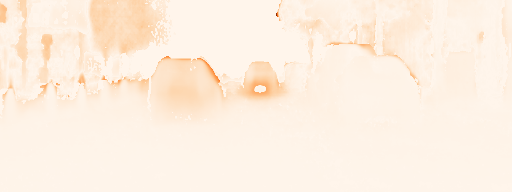}
		\end{minipage}\hspace{-0.5mm}
		\begin{minipage}[t]{0.192\textwidth}
			\centering
			\includegraphics[width=1\textwidth]{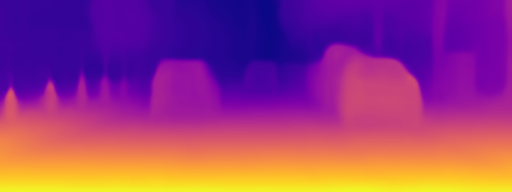}
		\end{minipage}\hspace{-0.5mm}
		\begin{minipage}[t]{0.192\textwidth}
			\centering
			\includegraphics[width=1\textwidth]{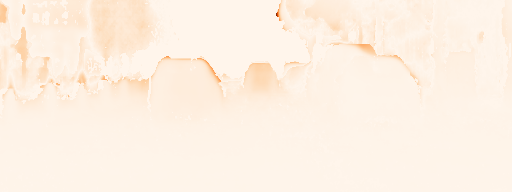}
		\end{minipage}
	}\\\vspace{-3mm}
	
	\centering
	\subfloat{
	    \begin{minipage}[t]{0.192\textwidth}
			\centering
			\includegraphics[width=1\textwidth]{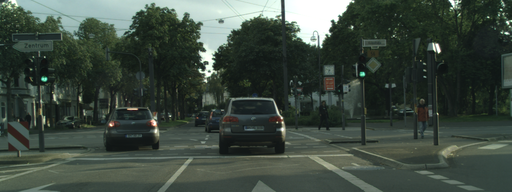}
		\end{minipage}\hspace{-0.5mm}
		\begin{minipage}[t]{0.192\textwidth}
			\centering
			\includegraphics[width=1\textwidth]{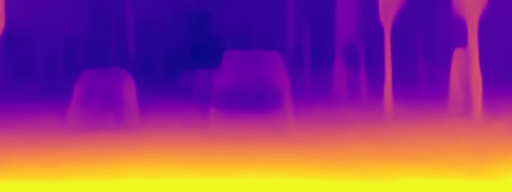}
		\end{minipage}\hspace{-0.5mm}
		\begin{minipage}[t]{0.192\textwidth}
			\centering
			\includegraphics[width=1\textwidth]{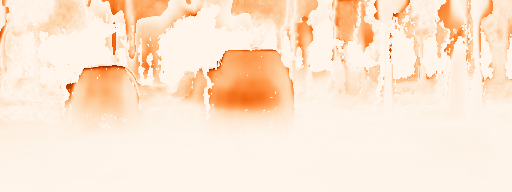}
		\end{minipage}\hspace{-0.5mm}
		\begin{minipage}[t]{0.192\textwidth}
			\centering
			\includegraphics[width=1\textwidth]{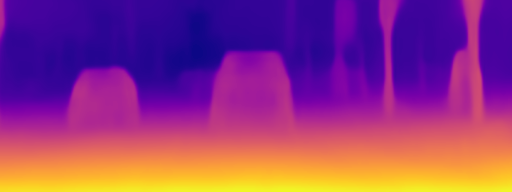}
		\end{minipage}\hspace{-0.5mm}
		\begin{minipage}[t]{0.192\textwidth}
			\centering
			\includegraphics[width=1\textwidth]{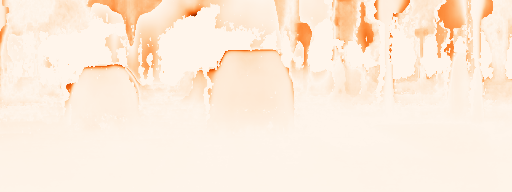}
		\end{minipage}
	}\\\vspace{-3mm}
	\centering
	\subfloat{
	    \begin{minipage}[t]{0.192\textwidth}
			\centering
			\includegraphics[width=1\textwidth]{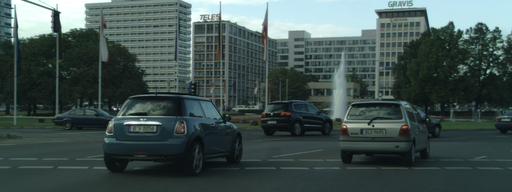}
		\end{minipage}\hspace{-0.5mm}
		\begin{minipage}[t]{0.192\textwidth}
			\centering
			\includegraphics[width=1\textwidth]{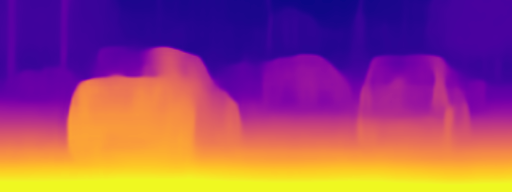}
		\end{minipage}\hspace{-0.5mm}
		\begin{minipage}[t]{0.192\textwidth}
			\centering
			\includegraphics[width=1\textwidth]{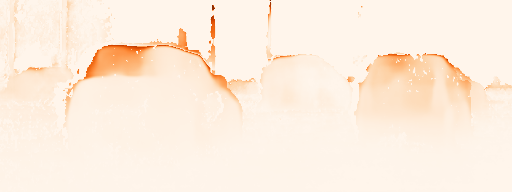}
		\end{minipage}\hspace{-0.5mm}
		\begin{minipage}[t]{0.192\textwidth}
			\centering
			\includegraphics[width=1\textwidth]{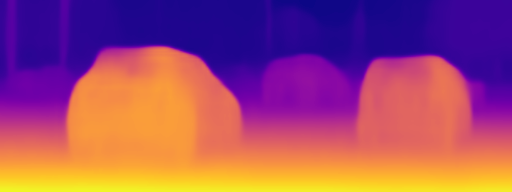}
		\end{minipage}\hspace{-0.5mm}
		\begin{minipage}[t]{0.192\textwidth}
			\centering
			\includegraphics[width=1\textwidth]{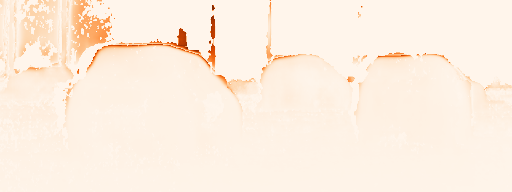}
		\end{minipage}
	}

    \caption{\textbf{The qualitative results} on Cityscapes dataset. Here, the 3 and 5 columns show the Abs.Rel.error, and the 2 and 4 columns show the predicted depth maps. It can be observed from the error map that our method significantly outperforms our baseline, especially in dynamic regions.}
    \label{fig:7}
\end{figure*}

\begin{figure*}
	\centering
    \rotatebox{90}{~~RGB}\hspace{-0.4mm}
	\subfloat{
	    \begin{minipage}[t]{0.19\textwidth}
			\centering
			\includegraphics[width=1\textwidth]{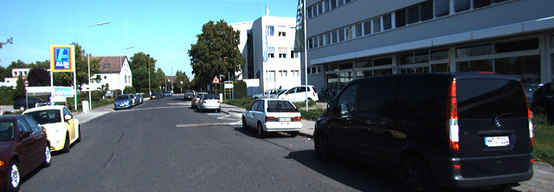}
		\end{minipage}\hspace{-0.5mm}
		\begin{minipage}[t]{0.19\textwidth}
			\centering
			\includegraphics[width=1\textwidth]{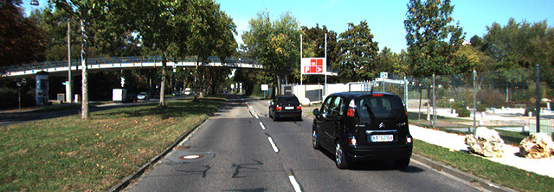}
		\end{minipage}\hspace{-0.5mm}
		\begin{minipage}[t]{0.19\textwidth}
			\centering
			\includegraphics[width=1\textwidth]{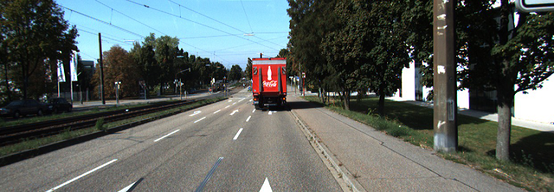}
		\end{minipage}\hspace{-0.5mm}
		\begin{minipage}[t]{0.19\textwidth}
			\centering
			\includegraphics[width=1\textwidth]{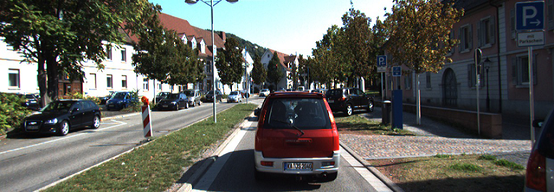}
		\end{minipage}\hspace{-0.5mm}
		\begin{minipage}[t]{0.19\textwidth}
			\centering
			\includegraphics[width=1\textwidth]{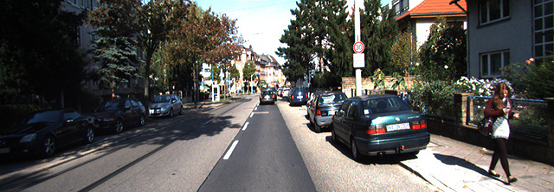}
		\end{minipage}
	}\\
	\vspace{-3mm}
	\centering
    \rotatebox{90}{~~~MD}\hspace{-0.4mm}
	\subfloat{
	    \begin{minipage}[t]{0.19\textwidth}
			\centering
			\includegraphics[width=1\textwidth]{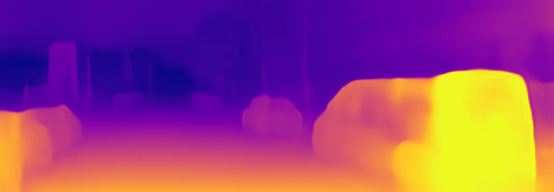}
		\end{minipage}\hspace{-0.5mm}
		\begin{minipage}[t]{0.19\textwidth}
			\centering
			\includegraphics[width=1\textwidth]{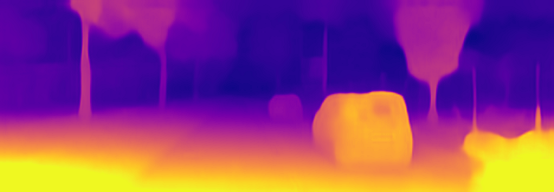}
		\end{minipage}\hspace{-0.5mm}
		\begin{minipage}[t]{0.19\textwidth}
			\centering
			\includegraphics[width=1\textwidth]{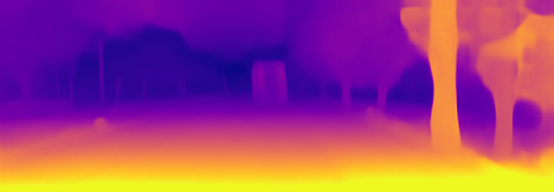}
		\end{minipage}\hspace{-0.5mm}
		\begin{minipage}[t]{0.19\textwidth}
			\centering
			\includegraphics[width=1\textwidth]{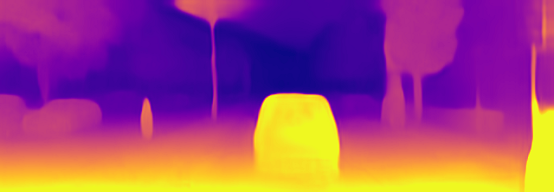}
		\end{minipage}\hspace{-0.5mm}
		\begin{minipage}[t]{0.19\textwidth}
			\centering
			\includegraphics[width=1\textwidth]{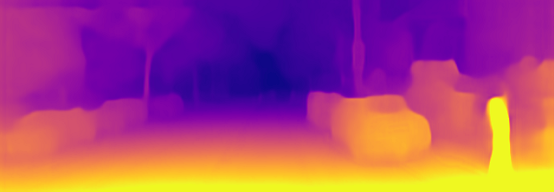}
		\end{minipage}
	}\\
	\vspace{-3mm}
	\centering
    \rotatebox{90}{~~~Our}\hspace{-0.4mm}
	\subfloat{
	    \begin{minipage}[t]{0.19\textwidth}
			\centering
			\includegraphics[width=1\textwidth]{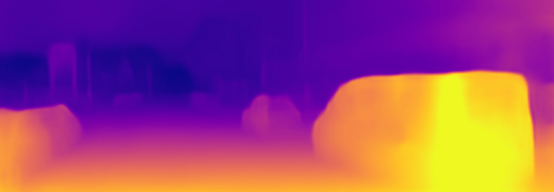}
		\end{minipage}\hspace{-0.5mm}
		\begin{minipage}[t]{0.19\textwidth}
			\centering
			\includegraphics[width=1\textwidth]{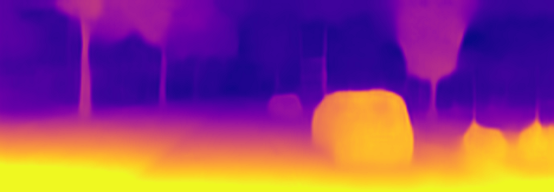}
		\end{minipage}\hspace{-0.5mm}
		\begin{minipage}[t]{0.19\textwidth}
			\centering
			\includegraphics[width=1\textwidth]{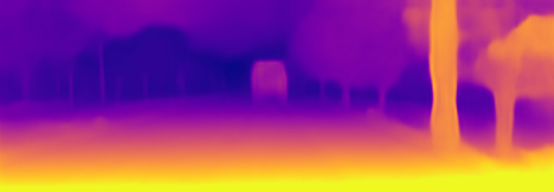}
		\end{minipage}\hspace{-0.5mm}
		\begin{minipage}[t]{0.19\textwidth}
			\centering
			\includegraphics[width=1\textwidth]{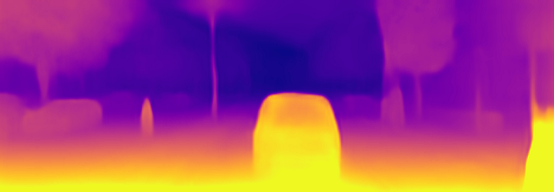}
		\end{minipage}\hspace{-0.5mm}
		\begin{minipage}[t]{0.19\textwidth}
			\centering
			\includegraphics[width=1\textwidth]{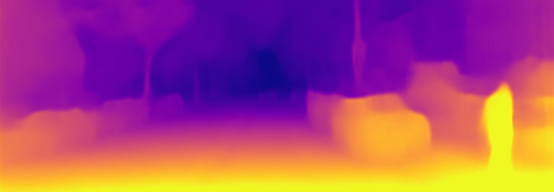}
		\end{minipage}
	}

    \caption{\textbf{The qualitative results} on KITTI dataset. Here we show our prediction results and our baseline (MD) \cite{manydepth} results. Since we focus on dynamic objects, we can see from the figure that our prediction results have better contour and texture information in dynamic regions. Furthermore, the corresponding qualitative results show in \Cref{fig:6}. }
    \label{fig:5}
\end{figure*}

\begin{figure*}
	\centering
	\subfloat{
	    \begin{minipage}[t]{0.19\textwidth}
			\centering
			\includegraphics[width=1\textwidth]{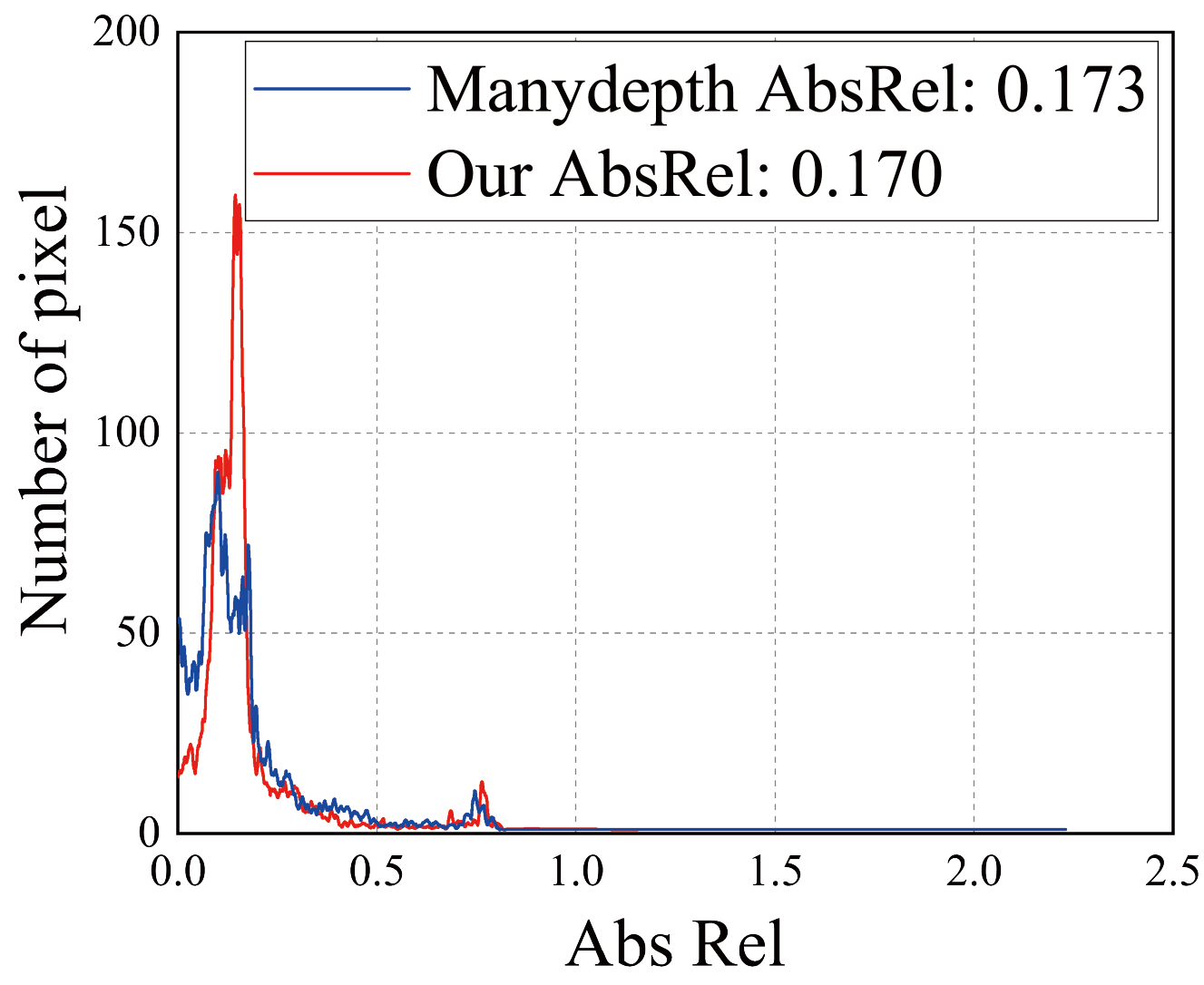}
		\end{minipage}\hspace{-0.5mm}
		\begin{minipage}[t]{0.19\textwidth}
			\centering
			\includegraphics[width=1\textwidth]{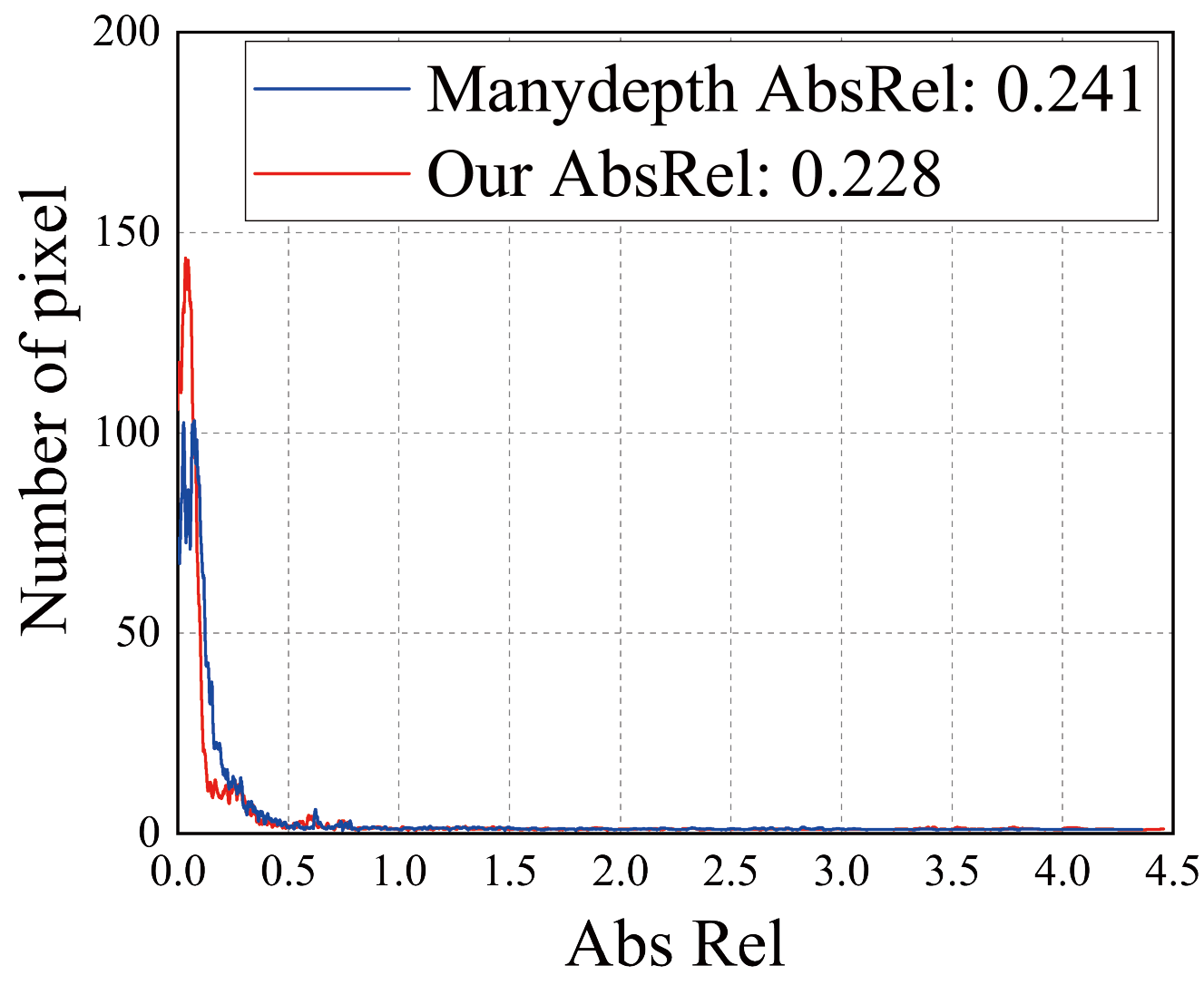}
		\end{minipage}\hspace{-0.5mm}
		\begin{minipage}[t]{0.19\textwidth}
			\centering
			\includegraphics[width=1\textwidth]{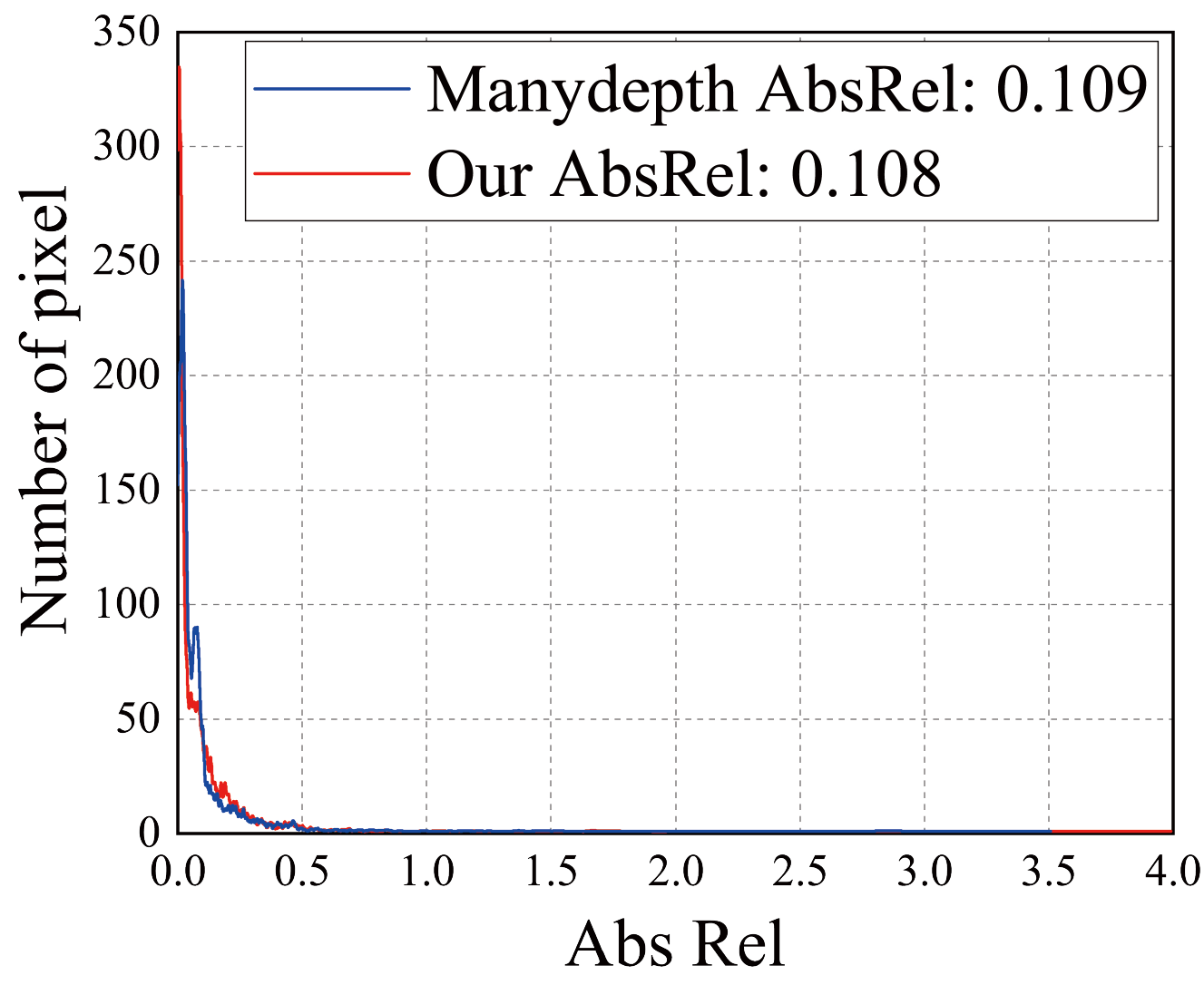}
		\end{minipage}\hspace{-0.5mm}
		\begin{minipage}[t]{0.19\textwidth}
			\centering
			\includegraphics[width=1\textwidth]{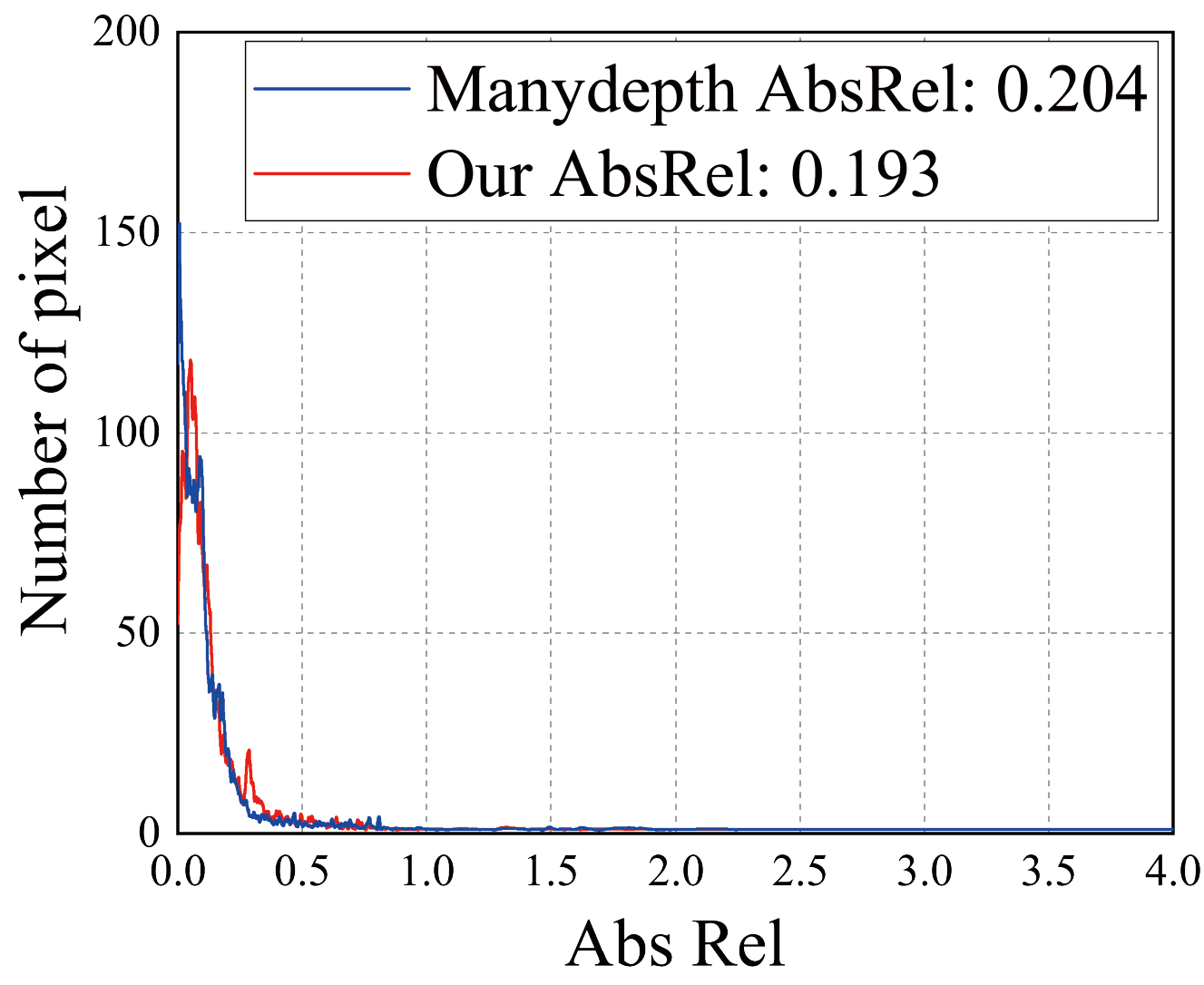}
		\end{minipage}\hspace{-0.5mm}
		\begin{minipage}[t]{0.19\textwidth}
			\centering
			\includegraphics[width=1\textwidth]{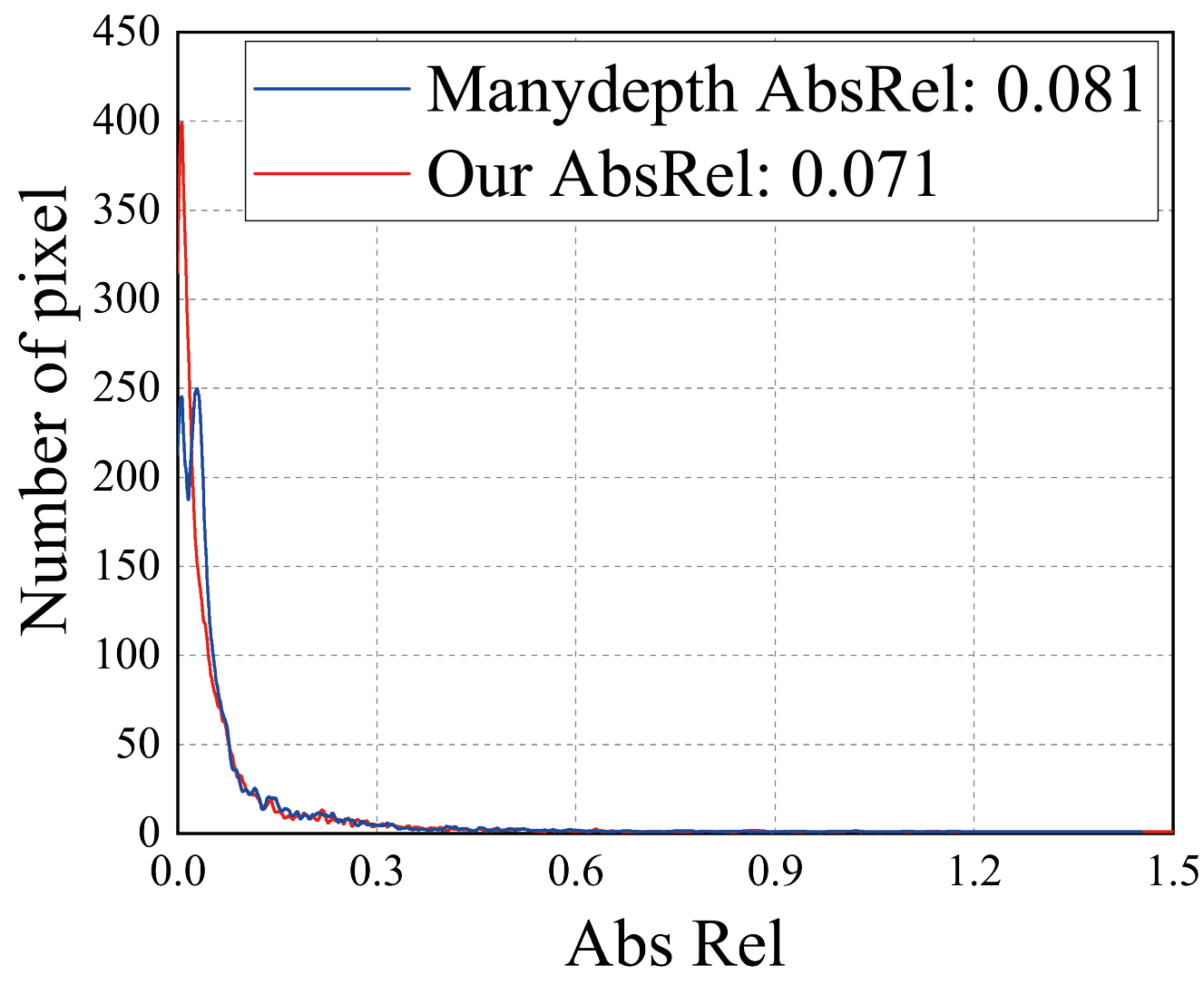}
		\end{minipage}
	}\\

    \caption{\textbf{The quantitative results} on KITTI dataset. Here we show the Abs.Rel. error statistics for each valid pixel of our method and baseline final depth map corresponding to \Cref{fig:5}.}
    \label{fig:6}
\end{figure*}

\subsubsection{Results on Cityscapes}
Below the \Cref{tab:1} shows each method score in the Cityscapes dataset, this dataset contains a large number of dynamic scenes. Our method currently outperforms all methods and compare to our baseline, the performance of our model achieves 12.3\% improvement. Moreover, \Cref{fig:8-a} provides consistency error (The difference between the depth of teacher network predictions and the lowest disparity map.), we get this error with the same parameters trained for 10 epochs and without any pretrained model, which improves by 7.43\% compared to the baseline. In addition, in \Cref{tab:7} we show the evaluation of the depth only for the dynamic region, our method improves 24.85\% in the dynamic region compared to the baseline.

\begin{table*}
\centering
\caption{\textbf{The depth evaluation results of the dynamic objects} (e.g. vehicles, bikes, and pedestrians) on the Cityscapes dataset. Dynamic object masks are generated by the pre-trained model EffcientPS \cite{mohan2021efficientps}.}
\label{tab:7}
\begin{tabular}{ccccccccc}
\toprule[1pt]
\multirow{2}{*}{Method} & \multirow{2}{*}{H x W} & \multicolumn{4}{c}{The lower the better} & \multicolumn{3}{c}{The higher the better}              \\ \cline{3-9} 
                        &                          & AbsRel   & SqRel   & RMSE    & RMSElog   & $\delta < 1.25$ & $\delta < 1.25^2$ & $\delta < 1.25^3$ \\ \hline
Monodepth2 \cite{monodepth2} & 416 x 128 & 0.159 & 1.937 & 6.363 & 0.201 & 0.816 & 0.950 & 0.981\\
InstaDM \cite{lee2021learning} & 832 x 256 & 0.139 & 1.698 &  5.760 & 0.181 & \textbf{0.859} & 0.959 & 0.982\\
Manydepth \cite{manydepth} & 416 x 128 & 0.169 & 2.175 & 6.634 & 0.218 & 0.789 & 0.921 & 0.969\\ \hline
Our (W/o PD Loss) & 416 x 128 & 0.130 & 1.163 & 5.953 & 0.183 & 0.801 & 0.955 & 0.986 \\
Our (W PD Loss) & 416 x 128 & \textbf{0.127} & \textbf{1.047} & \textbf{5.604} & \textbf{0.179} & 0.827 & \textbf{0.960} & \textbf{0.988} \\\bottomrule[1pt]
\end{tabular}
\end{table*}

\begin{table*}
\centering
\caption{\textbf{The results of using different backbones} for our optical flow module. KITTI${}^*$ is KITTI dataset with improved ground truth.}
\label{tab:8}
\begin{tabular}{ccccccccc}
\toprule[1pt]
\multirow{2}{*}{Method} & \multirow{2}{*}{Dataset} & \multicolumn{4}{c}{The lower the better} & \multicolumn{3}{c}{The higher the better}              \\ \cline{3-9} 
                        &                          & AbsRel   & SqRel   & RMSE    & RMSElog   & $\delta < 1.25$ & $\delta < 1.25^2$ & $\delta < 1.25^3$ \\ \hline
Our                     & KITTI                    & \textbf{0.095}    & \textbf{0.698}   & 4.329   & \textbf{0.173}     & \textbf{0.905}           & \textbf{0.966}             & \textbf{0.984}             \\
Our (ResNet18)          & KITTI                    & \textbf{0.095}    & 0.705   & \textbf{4.326}   & 0.174     & \textbf{0.905}           & \textbf{0.966}             & 0.983             \\ \hline
Our                     & KITTI${}^*$                   & \textbf{0.067}    & \textbf{0.359}   & 3.314   & \textbf{0.109}     & \textbf{0.946}           & \textbf{0.989}             & \textbf{0.997}             \\
Our (ResNet18)          & KITTI${}^*$                   & 0.068    & 0.360   & \textbf{3.284}   & \textbf{0.109}     & \textbf{0.946}           & \textbf{0.989}             & \textbf{0.997}             \\ \hline
Our                     & Cityspaces               & \textbf{0.100}    & \textbf{1.055}   & \textbf{5.884}   & \textbf{0.155}     & \textbf{0.899}           & \textbf{0.974}             & \textbf{0.991}             \\
Our (ResNet18)          & Cityspaces               & 0.102    & 1.129   & 5.961   & 0.157     & \textbf{0.899}           & 0.973             & 0.990             \\ \bottomrule[1pt]
\end{tabular}
\end{table*}

\begin{figure}[!ht]
    \centering
    \includegraphics[width=1\linewidth]{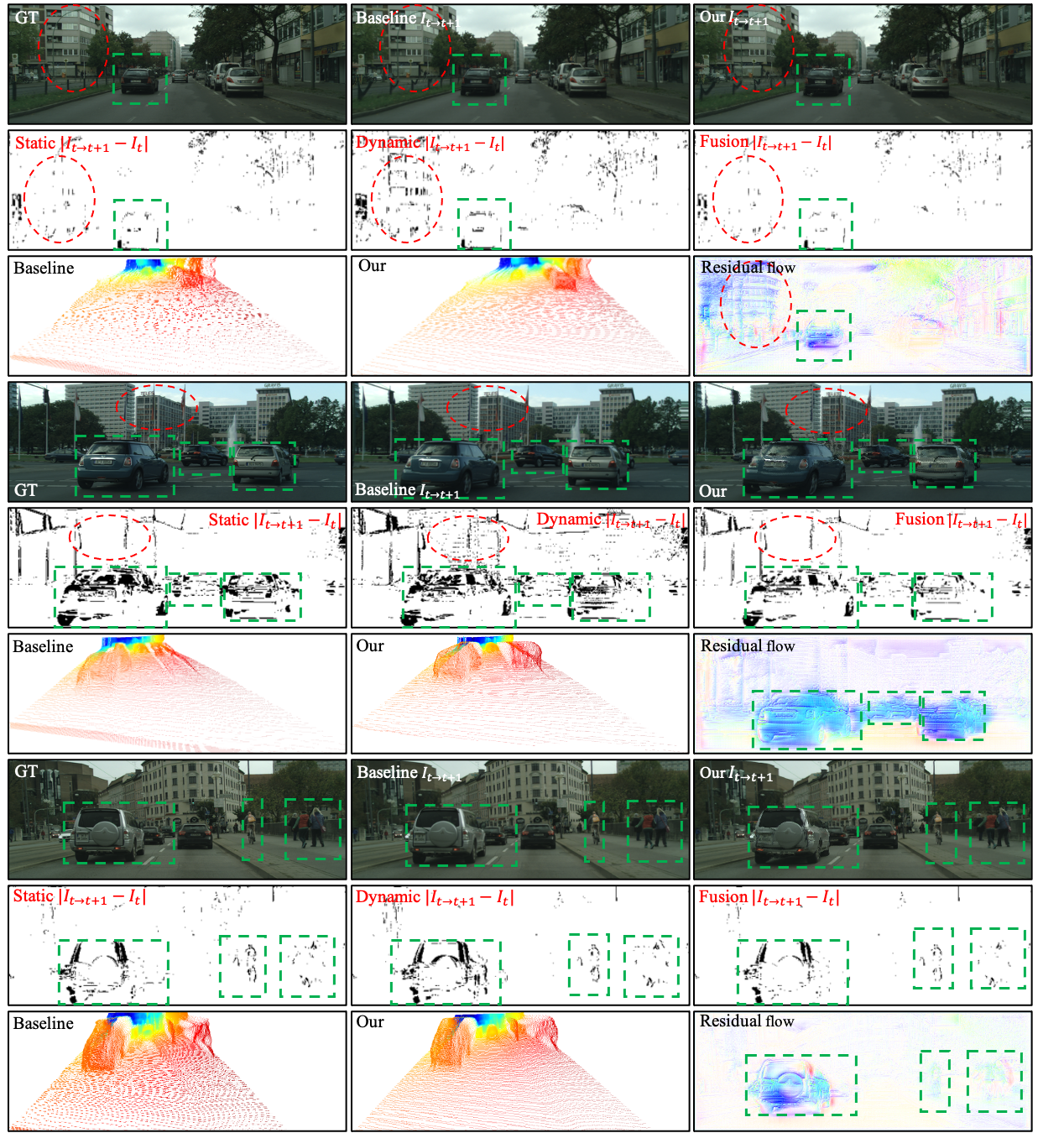}
    \caption{\textbf{The qualitative results} on Cityscapes. We provide the view synthesis, residual flow, occlusion map, and depth map converted to point cloud results.}
    \label{fig:12}
\end{figure}

\subsection{Qualitative Results}
The qualitative results are reported in \Cref{fig:5} and \Cref{fig:7}. In \Cref{fig:5}, our method performance is better in dynamic regions. We can see that the vehicles in columns 1, 3, and 4, the stones in column 2, and the people in column 5 have clearer texture, and contours information compared to the baselines.

As shown in \Cref{fig:7}, the depth predicted by our model significantly outperforms our baseline method, especially in dynamic regions, and the vehicles of our method are clearer and have no mismatched regions. The results, from the error map, indicate that the error of our method is smaller than that of our baseline method in the dynamic regions.

\begin{figure*}
\centering
\includegraphics[width=1\linewidth]{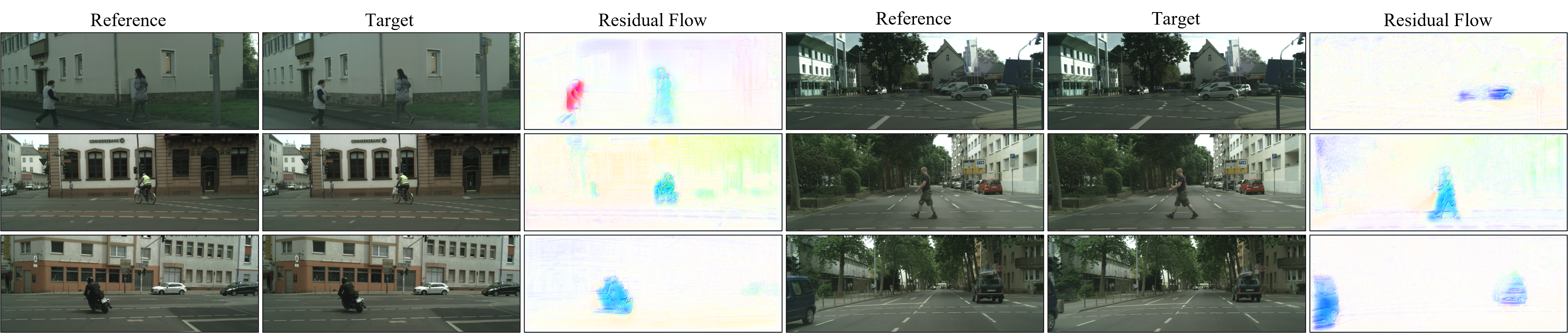}
\caption{\textbf{The qualitative result of the residual flow on Cityscapes dataset.} It is obvious that our optical flow network has captured moving objects.}
\label{fig:13}
\end{figure*}

In addition, in \Cref{fig:12}, only using dynamic will bring more occlusion and noise, while the fusion module can effectively reduce the occlusion, compared to our baseline. The residual flow results demonstrate our model's ability to accurately capture moving objects and handle inaccurate pixels without the use of priors, such as pre-trained segmentation models. Notably, the point cloud image converted from the depth map shows that our model performs better on moving objects. In \Cref{fig:13}, we provide more optical flow visualization results.

\subsection{Ablation Study}
In \Cref{tab:2} and \Cref{tab:5}, we provide an analysis of different components in our DS-Depth architecture on KITTI and Cityscapes, we validate the effectiveness of dynamic cost volume, three different forms of fusion modules, and pyramid distillation loss. In \Cref{tab:8}, we provide the results of the optical flow network using different backbones.

\subsubsection{Only Using Dynamic Cost Volume} In this setting, we only use the dynamic cost volume to train our model. In KITTI, the Abs.Rel. error does not change significantly, which is as we expected. Because the dynamic cost volume refines part of the occlusion areas but the same brings some extra occlusions and noise and the low number of dynamic objects in the KITTI dataset. As shown in \Cref{fig:8-b}, in the occlusion region, only using dynamic cost volume will degrade the performance of the model, and the predicted depth has already deviated from the ground truth. While in Cityscapes, the Abs.Rel. error improvement is significant, with a 4.39\% improvement. This improvement over KITTI is very significant, as a large number of moving objects are included in Cityscapes, and our residual flow succeeds in warping the moving objects to relatively correct positions.\\

\begin{table}[!t]
\caption{\textbf{Evaluate our method, L1, and SSIM} on KITTI and Cityscapes dataset to build cost volume.}
\label{tab:3}
\begin{tabular}{cccc|ccc}
\toprule[1pt]
\multirow{2}{*}{Method} & \multicolumn{3}{c|}{KITTI} & \multicolumn{3}{c}{Cityscapes} \\ \cline{2-7} 
                        & AbsRel  & SqRel  & RMSE   & AbsRel    & SqRel    & RMSE    \\ \hline
L1                      & 0.099   & 0.756  & 4.460  & 0.102     & 1.133    & 5.985   \\
SSIM                    & 0.099   & 0.778  & 4.507  & 0.102     & 1.136    & 5.964   \\
\textbf{Our}                     & \textbf{0.095}   & \textbf{0.698}  & \textbf{4.329}  & \textbf{0.100}     & \textbf{1.055}    & \textbf{5.884}   \\ \bottomrule[1pt]
\end{tabular}
\end{table}

\subsubsection{Two Cost Volumes with Fusion Module} 
\begin{figure}[t]
  \centering
  \subfloat[\label{fig:8-a}]{
  \includegraphics[width=0.46\linewidth]{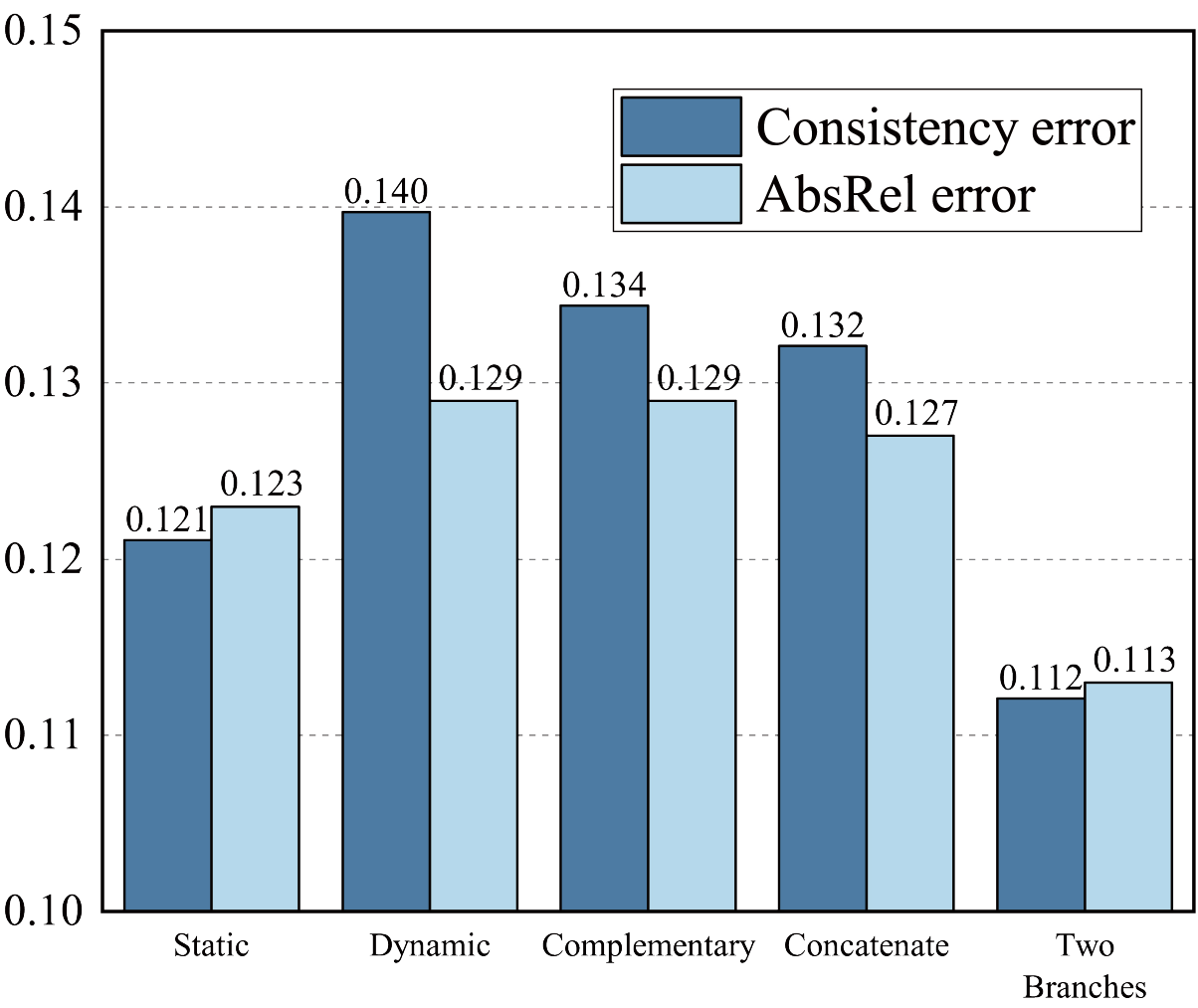}
    }
  \subfloat[\label{fig:8-b}]{
\includegraphics[width=0.46\linewidth]{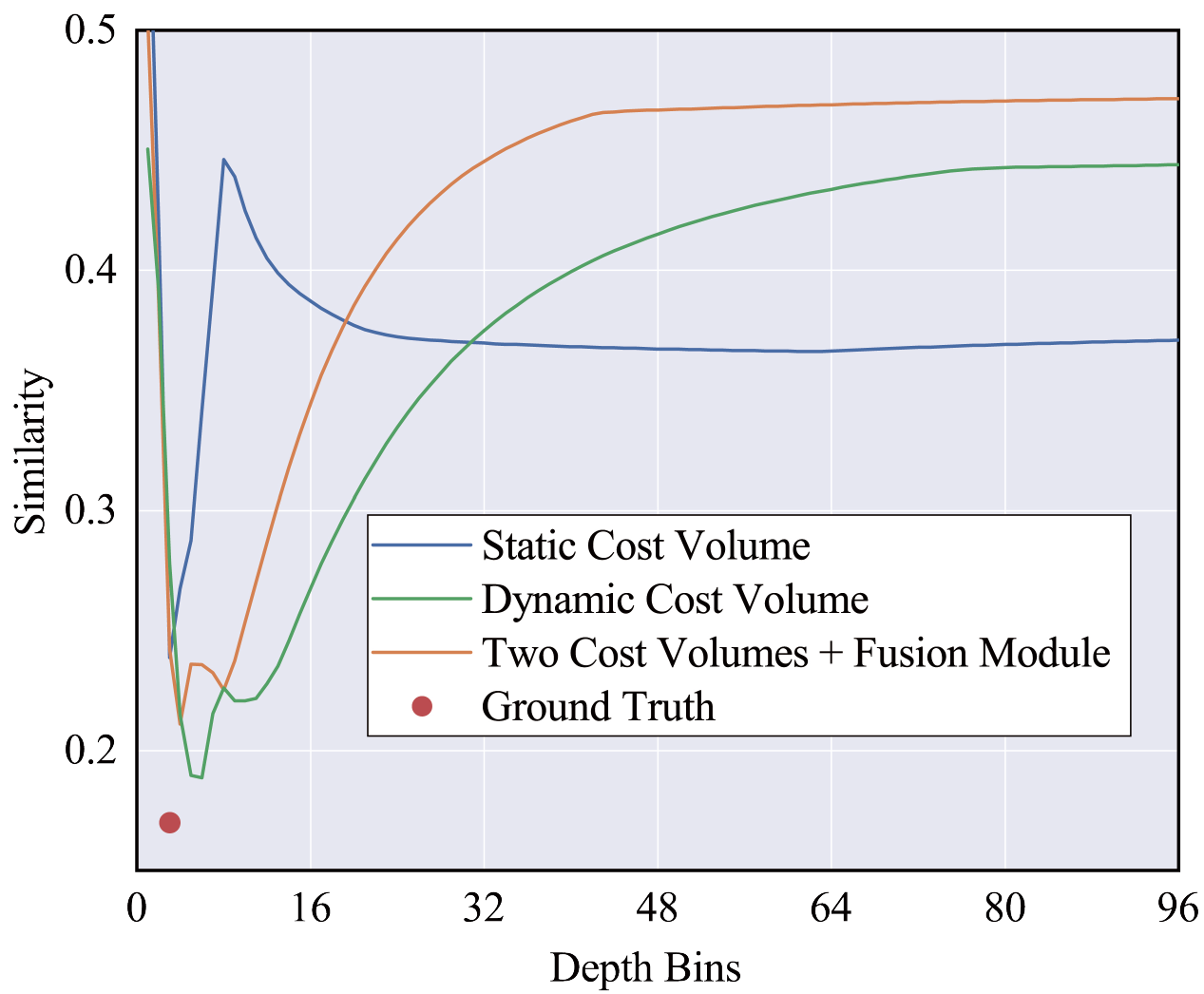}
    }
  \caption{(a) Consistency error and Abs.Rel. error between a teacher network and the lowest cost volume trained for 10 epochs with no pretrained model in the Cityscape dataset. (b) The matching probability along depth bins in the occlusion area. The blue line is our baseline, the orange line only uses dynamic cost volume and the green line uses the fusion module.}
  \label{fig:8}
\end{figure}

We tried three fusion modules 1) complementary, 2) simple concatenate, and 3) two-branch. In KITTI, compared to only using the dynamic cost volume, the fusion module increases the performance of the model (0.9\%-6.86\%) and also alleviates the occlusion problem to a certain extent. Our results on Cityscapes also support this conclusion (4.59\%-6.42\%). \Cref{fig:8-a} illustrates that if the dynamic cost volume only is used, the performance of our model will decrease due to the extra occlusion areas and noise. When we leverage the fusion module, the error is significantly reduced and this suggests that our fusion module alleviates this problem to a certain extent, please see the effect of our model in the bottom half of \Cref{fig:0}. In \Cref{fig:8-b}, when using the fusion module, the predicted depths are closer to the ground truth than using only the dynamic cost volume, which confirms that our fusion layer is effective in occluded regions. After our observation, it became evident that the complementary module's functionality on the KITTI dataset is rather limited. This arises from the dataset's distinctive trait of having a lower occurrence of moving objects, with a significant majority of them being stationary. As a result, the effectiveness of the complementary module appears to be diminished when applied to the KITTI dataset. This is because there are two cases in the complementary module: one of the same region in the dynamic and static cost volume is occluded, then the module will select the unoccluded cost volume, and if both cost volumes are unoccluded in the same region then the module will select the cost volume with the lower error. However, our cost volume is composed jointly by L1 and ssim errors, so there are cases of mismatching in some regions. If the error of dynamic cost volume is small and there is a lot of noise, and our module incorrectly selects this part of the dynamic cost volume, then the situation shown in our ablation experiment will occur. In contrast, for the concatenate module, because this module is learnable, the above problem does not exist.

Moreover, the effect of the two-branch fusion module is better than the complementary and simply concatenate fusion module, because the complementary fusion module obtains a small error but some regions are greatly affected by artifacts. For the simple concatenate fusion module, it cannot directly get the most correct complementary error, hence the effect is not significantly improved.

\begin{figure}[!ht]
\centering
\includegraphics[width=1\linewidth]{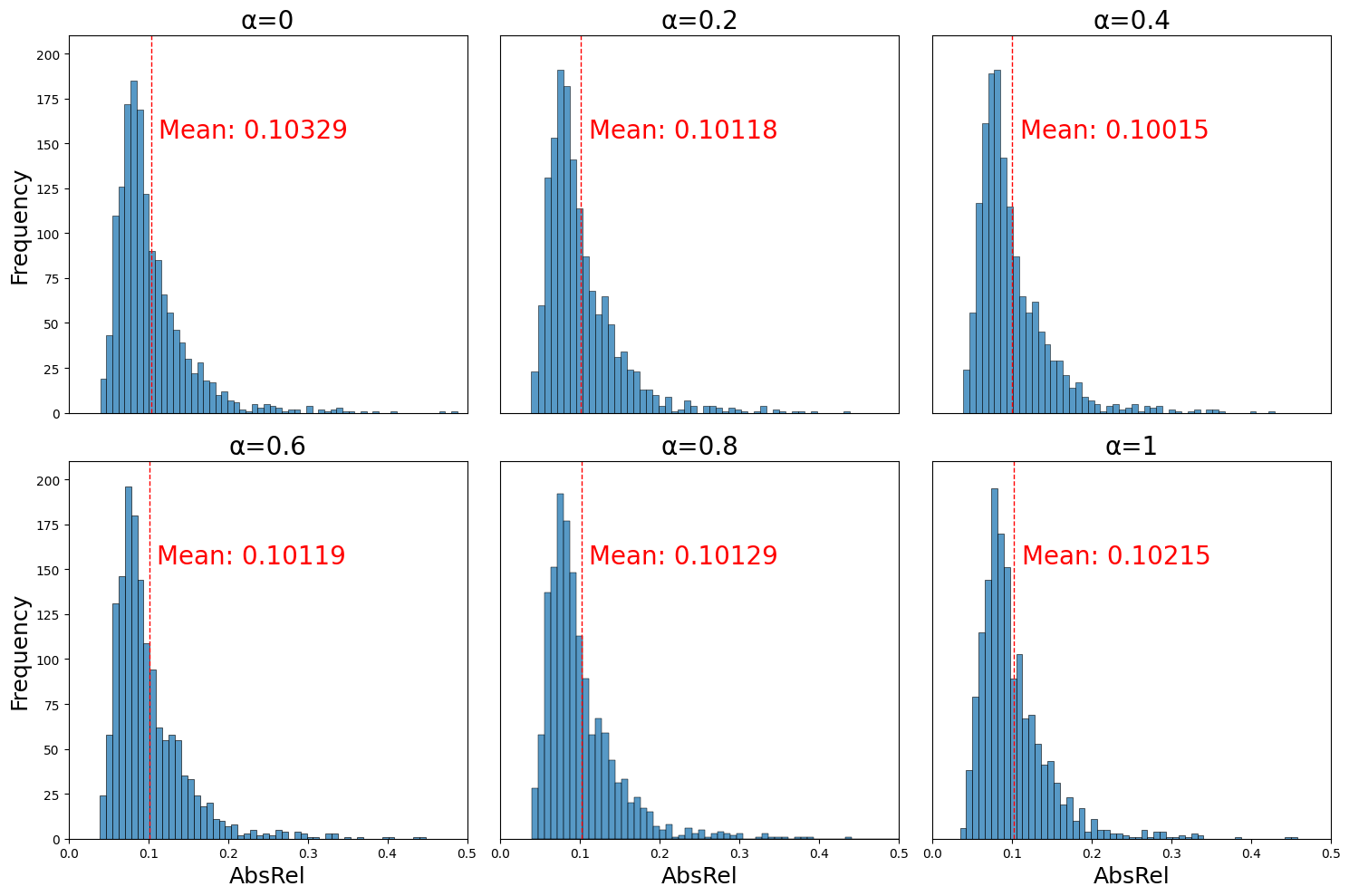}
\caption{\textbf{The comparison of the cost volume constructed} by different $\alpha$ for the performance of the model in the Cityscape dataset. When $\alpha=0.4$, the model performance is the best, and the Abs.Rel. error distribution is relatively smoother.}
\label{fig:9}
\end{figure}

\subsubsection{Pyramid Distillation Loss}
The main purpose of this loss is to fix the inaccurate photometric error caused by low resolution. Based on our observations, the photometric error of our model is only less accurate at 1/8 resolution, this component has little effect on the performance of our model. However, as shown in \Cref{tab:7}, the effect of this loss is relatively obvious in dynamic regions.

In addition, we also provide three different ways to build cost volume results.  Although constructing the cost volume in the L1 way can alleviate the dynamic problem to a certain extent, this way sacrifices the surface details, while SSIM can provide more surface details, thus we try to construct the cost volume in the way of photometric error. \Cref{tab:3} presents to construct cost volume using the photometric error method, which is better than L1 and SSIM. As shown in \Cref{fig:9}, we use \Cref{eq:18}, we explore the impact of different $\alpha$ on model performance on cityscapes. Among them, when $\alpha=0.4$, the model has the best performance and has a greater improvement compared to other hyperparameters.

\section{Conclusion}
In this work, we proposed DS-Depth, a general self-supervised depth estimation model framework. Specifically, we proposed a dynamic cost volume construction by combining camera ego-motion and residual optical flow to optimize the cost volume occlusion problem in dynamic regions. To alleviate the extra occlusion and noise caused by dynamic cost volume, the adaptive fusion module is designed to effectively improve the contour and texture information of dynamic objects. Moreover, we proposed a pyramid distillation loss to address the inaccuracy of photometric error at low resolutions and an adaptive photometric error loss to alleviate large gradients in the occlusion region. During our experiments, we found that the accuracy of the single-frame teacher network will have a great impact on the performance of the model in the later stage of training, especially on the Cityscapes dataset, which is very obvious and will bring some negative effects. Therefore, exploring how to optimize single-frame networks may further improve the performance of multi-frame methods.

\section*{Acknowledgments}
This work is supported by the UK Medical Research Council (MRC) Innovation Fellowship under Grant MR/S003916/2, International Exchanges 2022 IEC$\backslash$NSFC$\backslash$223523 and Securing the Energy/Transport Interface EP/X037401/1.



 
\bibliographystyle{ieeetr}
\bibliography{myref}

\end{document}